%% file: main.tex
\PassOptionsToPackage{table}{xcolor}
\PassOptionsToPackage{capitalise}{cleveref}
\documentclass{fairmeta}

\usepackage{latexsym}
\usepackage[utf8]{inputenc}
\usepackage{amsmath,amssymb}
\usepackage{etoc}
\usepackage{float}%
\usepackage{wrapfig}%

\makeatletter
\DeclareRobustCommand{\levsc}{L{\fontsize{\strip@pt\dimexpr.8\dimexpr\f@size pt\relax\relax}{\f@baselineskip}\selectfont EV}}
\makeatother
\newcommand{\name}{\texorpdfstring{\levsc}{Lev}}
\newcommand{\levbench}{LevBench}
\newcommand{\kevseen}{Kev-Seen}
\newcommand{\kevtransfer}{Kev-Transfer}
\newcommand{\kevskills}{Kev-Skills}
\newcommand{\kevbreadth}{Kev-Breadth}
\newcommand{\trainsmall}{Train-31k}
\newcommand{\trainmid}{Train-60k}
\newcommand{\trainlarge}{Train-120k}
\definecolor{loopblue}{HTML}{1F5AA6}
\definecolor{rowgray}{HTML}{F2F2F2}
\newcommand{\ci}[1]{{\scriptsize\color{black!55}#1}}
\newtcolorbox{takeaway}{colback=white, colframe=black, boxrule=0.5pt, sharp corners,
  left=6pt, right=6pt, top=4pt, bottom=4pt, before skip=10pt, after skip=10pt, fontupper=\small\raggedright}

\newcommand{\colfig}{0.6\textwidth}

\title{\name: Thinking Fast and Deep with Looped Decision Models}

\author[1,*]{Dingjie Song}
\author[1,*]{Zhiling Yan}
\author[2,3]{Yuxuan Zhang}
\author[1,\dagger]{Lichao Sun}

\affiliation[1]{Lehigh University}
\affiliation[2]{University of British Columbia}
\affiliation[3]{Vector Institute}

\contribution[*]{Equal contribution}
\contribution[\dagger]{Corresponding author}

\abstract{%
Decision models answer typed questions about an input with one probability per option, computed in a single forward pass without generating text.
Like the fast System 1 of dual-process theory, they answer at once, but their depth is fixed at training time, so they spend the same computation on easy and hard questions.
Models that reason in generated text can think longer, but they pay for every extra step in decoded tokens.
We present \name{}, a decision model that thinks deeper by rerunning its layers in a loop instead of generating tokens.
\name{} builds on a language model pretrained as a loop, reads the options after every loop step and trains all steps jointly, so one checkpoint can answer at any depth.
On three looped backbones, accuracy rises with depth on the same weights, mostly on hard questions, and a 2.6B \name{} goes from 63.2\% at depth one to 76.2\% at depth four on JevBench.
On-demand depth, set per question by a small stopper with a threshold fitted on unlabelled requests, saves 15\% to 24\% of the computation at a cost of under one accuracy point, level with or slightly above a fixed mix of depths.
On a locked test set, the 1.4B and 3B models are 5.4 to 8.8 points more accurate than dense (non-looped) models and cascades of the same latency below 50\,ms.%
}

\metadata[Code]{\url{https://github.com/OpenLAIR/lev}}
\metadata[Models]{\url{https://huggingface.co/collections/songdj/lev}}
\metadata[Website]{\url{https://openlair.github.io/lev/}}

\hypersetup{
  pdftitle={LEV: Thinking Fast and Deep with Looped Decision Models},
  pdfauthor={Dingjie Song, Zhiling Yan, Yuxuan Zhang, Lichao Sun}
}

\newcommand{\onelogo}[1]{\raisebox{-0.5\height}{\includegraphics[height=0.55cm]{logos/#1}}}
\renewcommand{\titlelogos}{\onelogo{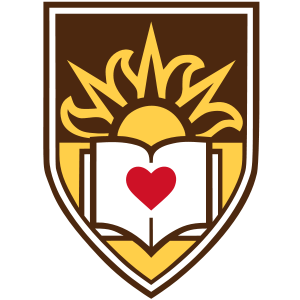}\;\onelogo{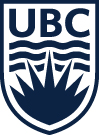}\;\onelogo{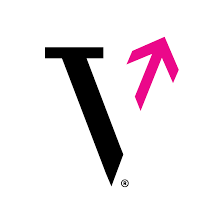}}

\begin{document}
\etocdepthtag.toc{mtmain}
\maketitle

\input{sections/intro}
\input{sections/related}
\input{sections/motivation}
\input{sections/method}
\input{sections/setup}
\input{sections/results}
\input{sections/analysis}
\input{sections/conclusion}
\input{sections/limitations}

\bibliographystyle{bibstyle}
\bibliography{custom}

\clearpage
\beginappendix
\etocdepthtag.toc{mtappendix}
\etocsettagdepth{mtmain}{none}
\etocsettagdepth{mtappendix}{subsection}
\etocsettocstyle{\section*{Appendix Contents}}{}
\tableofcontents
\let\appsection\section\renewcommand{\section}{\FloatBarrier\appsection}
\setcounter{topnumber}{3}\setcounter{dbltopnumber}{3}\setcounter{totalnumber}{5}
\renewcommand{\topfraction}{0.9}\renewcommand{\dbltopfraction}{0.93}\renewcommand{\textfraction}{0.08}
\renewcommand{\floatpagefraction}{0.8}\renewcommand{\dblfloatpagefraction}{0.8}

\input{sections/appendix}

\FloatBarrier

\end{document}

%% file: sections/intro.tex
\section{Introduction}
\label{sec:intro}

\begin{wrapfigure}[22]{r}{0.4\textwidth}
  \centering
  \vspace{-18pt}
  \captionsetup{font=footnotesize,aboveskip=3pt}
  \includegraphics[width=0.4\textwidth]{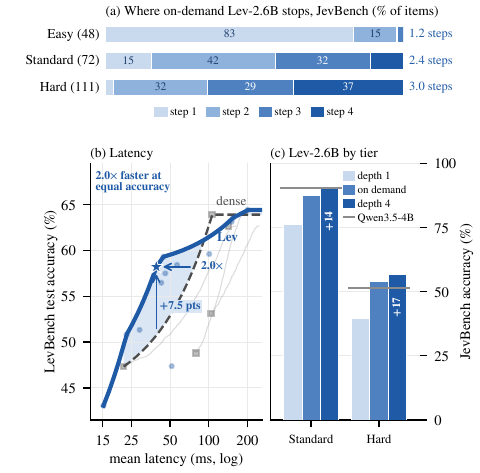}
  \caption{(a)~Loop steps that on-demand \name{}-2.6B spends per JevBench tier. (b)~\levbench{} test accuracy against latency, with the Pareto frontiers of \name{} over all depths (solid) and of dense models and cascades (dashed). Arrows mark the gap of on-demand \name{}-3B to the dense frontier. (c)~JevBench accuracy of \name{}-2.6B on the standard and hard tiers at depth one, on demand, and at depth four, against the dense Qwen3.5-4B control.}
  \label{fig:teaser}
\end{wrapfigure}
Generative language models that reason with chain-of-thought act as a System~2 \citep{kahneman2011thinking}.
They write out their reasoning before choosing an option, so every decision is slow, even an easy one \citep{villalon2026djev,razorback2026openjev}, and turning the reasoning off lowers their accuracy markedly on JevBench \citep{standhartinger2026jevbench}.
Decision models such as Jev \citep{typesafe2026jev} and Kev \citep{palmer2026kev} are a System~1 for questions with a fixed set of options.
They return a probability for each option from one forward pass \citep{devlin2019bert,robinson2023leveraging} in tens of milliseconds, but they spend the same computation on every question and cannot think longer on a hard one.
Not every question needs more computation.
On the public JevBench items, Kev-4B answers 28 more items correctly than Kev-0.8B, and \cref{fig:motivation}a shows that all of them come from the standard and hard tiers, because both models already answer every easy item.
A decision model that could add computation only where it pays off would keep the speed of the small model on easy questions and approach the accuracy of the large one on hard questions.

Existing ways to spend more computation without generating text fall short for decision models (\cref{sec:motivation}).
A cascade passes uncertain questions to a larger model \citep{varshney2022cascading,chen2024frugalgpt}, but reads them again and needs a second model.
Early exit stops at a confident intermediate layer \citep{xin2020deebert,schuster2022calm,elhoushi2024layerskip}, but these layers are not trained to be read and are often confidently wrong on new data \citep{meronen2024fixing}.
A loop reruns the same layers without adding parameters \citep{dehghani2019universal,geiping2025recurrent}, but added to a pretrained model it barely helps, as Jev-LCT \citep{lct2026} reports, unless it gets far more continued pretraining than a decision model is trained on \citep{mcleish2025retrofit,koishekenov2025etd}.

We propose \name{}, a System~1.5 decision model \citep{wang2025system15} that thinks in loops rather than tokens.
\name{} builds on natively looped language models \citep{zhu2025ouro,nanbeige2026}, whose loop is trained in pretraining rather than added afterwards.
Decision training therefore does not have to teach the model to use its loop, whereas a loop added to a pretrained decision model stays unused (\cref{sec:motivation}).
A shared readout scores the options after each step and all steps are trained jointly, so one set of weights answers at any depth, and a further step resumes from the previous hidden states without reading the input again.
A lightweight stopper sets the depth of each question on demand by predicting whether its answer already agrees with the deepest one, with a threshold set on unlabelled requests.
We use this target because the usual early-exit rule, a confidence threshold set on labelled data, does not beat a fixed mix of depths here, as shallow readouts are overconfident on new data.
Since the target is the model's own deepest answer, a deployment can fit the stopper on its own traffic without labels.

We build \name{} on three looped backbones and evaluate it on JevBench and two held-out benchmarks.
Accuracy rises with depth on the same weights, mostly on hard questions (\cref{fig:teaser}c), and the stopper spends more steps on harder questions without seeing their difficulty (\cref{fig:teaser}a).
On-demand depth saves 15\% to 24\% of the computation at a cost of under one accuracy point, level with or slightly above a fixed mix of depths.
A locked test, read once after the recipe and the stopper were fixed, confirms that on-demand depth stays within one point of the deepest readout.
Below 50\,ms, the 1.4B and 3B models are 5.4 to 8.8 points more accurate than dense (non-looped) models and cascades of the same latency (\cref{fig:teaser}b).

Our contributions are as follows.
\begin{itemize}
  \item We propose \name{}, a System~1.5 decision model that thinks deeper on hard questions through loop steps rather than generated text. It reads the options after every step of a natively looped backbone, so one checkpoint serves every depth.
  \item We show that a stopper whose threshold is set on unlabelled requests gives on-demand depth without labels. It predicts agreement with the deepest answer rather than correctness, and it is level with or above a fixed mix of depths where confidence thresholds set on labelled data fall below it.
  \item On three looped backbones, \name{} gains most from depth on hard questions, and below 50\,ms it is more accurate than dense models and cascades of the same latency. We also build \levbench{}, a benchmark frozen before any model read it, whose tiers move step by step away from the training data.
\end{itemize}

%% file: sections/related.tex
\section{Related Work}
\label{sec:related}

\paragraph{Decision models.}
Decision models score options discriminatively in one forward pass \citep{devlin2019bert,zaratiana2024gliner,robinson2023leveraging,zheng2024selection}, and block attention masks let independent questions share that pass \citep{ratner2023pcw}.
Jev and its open reimplementation Kev read the options with a pointer head \citep{vinyals2015pointer}.
A generative model can also answer such questions by naming an option, but it must bind each option to its symbol \citep{robinson2023leveraging} and is sensitive to the order and labels of the options \citep{zheng2024selection}.
Open Jev-style servers such as djev \citep{villalon2026djev} and OpenJev \citep{razorback2026openjev} follow this generative route and can add reasoning text before the answer, which raises accuracy on hard questions at the price of decoding time on every question.
A decision model instead scores the options themselves in one pass, and its probabilities can be calibrated with a single temperature \citep{guo2017calibration}.
\name{} adds per-question depth to such a model through a backbone pretrained as a loop.

\paragraph{Looped models and adaptive computation.}
Looped transformers rerun one block of layers to add computation without adding parameters \citep{dehghani2019universal,giannou2023looped}, and looped language models such as Ouro \citep{zhu2025ouro} reason better with more iterations and can be read at several depths \citep{geiping2025recurrent,saunshi2025latent,jeddi2026loopformer,popescu2026adaptive}.
A loop added to a standard pretrained model needs adapters or continued pretraining \citep{bae2024relaxed,mcleish2025retrofit} and otherwise tends to stay unused \citep{chen2026trainingfree,shapiro2026retrofitting}.
Adaptive computation stops early through halting \citep{graves2016act} or confident exits \citep{xin2020deebert,elbayad2020depth,schwartz2020right}, or adds latent tokens \citep{hao2024coconut,goyal2024pause}, and its shallow exits are often overconfident \citep{meronen2024fixing} unless calibrated toward the full model, as in CALM \citep{schuster2022calm}.
CATs \citep{schuster2021cats} likewise predicts whether an exit agrees with the full model, and BERxiT \citep{xin2021berxit} learns when to exit from labelled data.
\name{} instead trains every loop step as a decision model, stops whole steps without adding positions, and sets the threshold of its stopper on unlabelled requests.

%% file: sections/motivation.tex
\section{Motivation}
\label{sec:motivation}

\begin{figure*}[t]
  \centering
  \includegraphics[width=\textwidth]{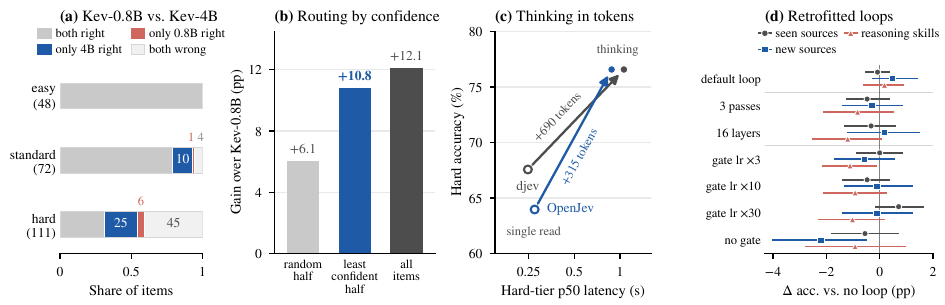}
  \caption{Evidence for depth chosen per question. (a)~Outcome of each public JevBench item under Kev-0.8B and Kev-4B. (b)~Gain over Kev-0.8B when Kev-4B answers a random half, the least confident half, or all items. (c)~Single read against generated reasoning on the same base model, with latencies from published results on different endpoints. (d)~Accuracy change from loops retrofitted onto Qwen3.5-0.8B, each row changing one setting of the default.}
  \label{fig:motivation}
\end{figure*}

A decision model spends the same computation on every question.
To test whether this is a good use of computation, we compare the released Kev-0.8B and Kev-4B, trained on the same sources, item by item on the 231 public JevBench items (\cref{app:motivation}), and we try three ways of adding computation.
They lead to two observations.

\paragraph{Observation 1: Extra computation pays off on few questions, and a cheap first read finds them.}
(1)~\textbf{The gain comes from hard questions.}
Kev-4B answers 28 more items correctly than Kev-0.8B, and \cref{fig:motivation}a shows that this gain comes entirely from the standard and hard tiers, because both models already answer every easy item.
(2)~\textbf{The small model can tell which questions need more.}
The two models reach different outcomes on only 42 of the 231 items, yet Kev-4B spends its extra parameters on every item.
The confidence of Kev-0.8B finds these items.
As \cref{fig:motivation}b shows, sending its least confident half to Kev-4B nearly matches Kev-4B on all items.

\paragraph{Observation 2: Existing ways to add computation cost too much or stay unused.}
(1)~\textbf{Cascades read uncertain questions twice.}
Kev-0.8B already takes 79\,ms at the median against 108\,ms for Kev-4B, so every item the cascade of \cref{fig:motivation}b passes on takes longer than Kev-4B alone, is read again from the start, and needs a second model to host.
(2)~\textbf{Thinking in tokens pays on every question.}
JevBench scores two open Jev-style servers, djev \citep{villalon2026djev} and OpenJev \citep{razorback2026openjev}, with and without generated reasoning on the same DiffusionGemma 26B base model.
These results in \cref{fig:motivation}c show that thinking raises hard accuracy by 9 to 13 points but makes the median hard item three to four times slower, and every question pays for its reasoning text, including easy ones.
(3)~\textbf{Retrofitted loops stay closed.}
\label{sec:retrofit}
We loop eight middle layers of Qwen3.5-0.8B-Base, the backbone of Kev-0.8B, behind per-channel gates that start at zero, and train the model with Kev's recipe (\cref{app:retrofit}).
\Cref{fig:motivation}d shows that the loop changes accuracy by less than half a point, and that a higher gate learning rate, a third pass, or a wider block does not help.
The gates stay near zero, so the loop stays closed, while removing the gates lowers accuracy on new sources by 2.2 points.
A backbone pretrained as a loop is open from the start.

%% file: sections/method.tex
\section{Method}
\label{sec:method}

\begin{figure*}[t]
  \centering
  \includegraphics[width=\textwidth]{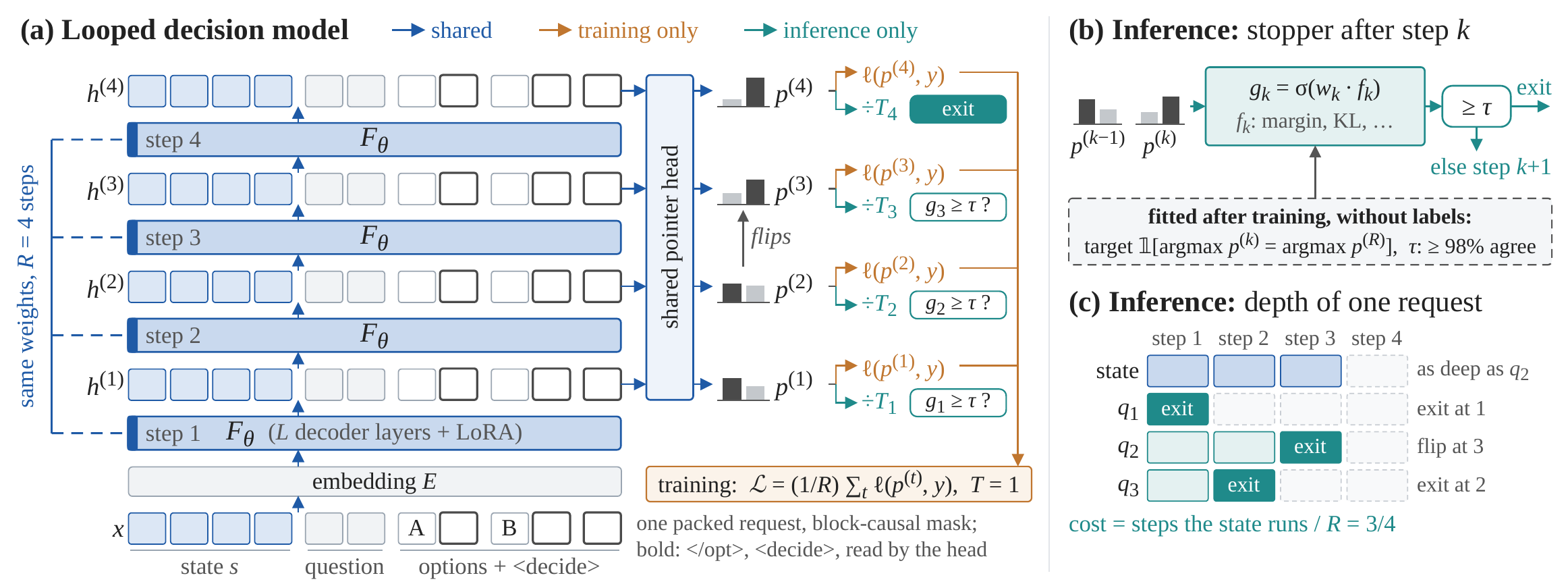}
  \caption{\name{} with $R$ loop steps, drawn for $R=4$. (a) A request is packed into one sequence, and the same decoder stack $F_\theta$ runs $R$ times. One shared pointer head reads the \texttt{</opt>} and \texttt{<decide>} positions after every step. (b) After step $k$, a stopper predicts whether the answer already agrees with the deepest one. (c) Each question exits at its own depth, and the state runs as deep as its deepest question.}
  \label{fig:method}
\end{figure*}

\subsection{Preliminaries}
\label{sec:prelim}
\paragraph{Decision models.}
A request holds a state $s$ and questions $q_1,\dots,q_m$, and question $q_j$ lists options $o_{j1},\dots,o_{jK_j}$.
Following Kev, the state comes first, and each question follows with its options wrapped in delimiter tokens and a final \texttt{<decide>} token.
A block-causal mask lets each question attend only to the state and to its own tokens, and its position indices restart after the state, so each question is answered as if it were asked alone while the state is encoded once.
With $h_{\mathrm{dec}}$ the final hidden state at \texttt{<decide>} and $h_k$ the one at the closing delimiter of option $k$, the pointer head computes
\begin{equation}
  z_k = \frac{(W_K h_k)^{\top} (W_Q h_{\mathrm{dec}})}{\sqrt{d_p}},
  \label{eq:pointer}
\end{equation}
where $W_Q, W_K \in \mathbb{R}^{d_p \times d}$ are learned, and returns $p = \mathrm{softmax}(z / T)$ with a temperature $T$ fitted after training.
Training minimises log loss $\ell(p, y) = -\log p_y$ at $T=1$, or cross-entropy for the few questions whose label $y$ is a distribution.

\paragraph{Looped transformers.}
A looped transformer applies one stack of $L$ decoder layers $F_\theta$ several times with the same weights \citep{dehghani2019universal,saunshi2025latent}.
From the token embeddings $h^{(0)} = E(x)$, step $t$ computes
\begin{equation}
  h^{(t)} = \mathrm{Norm}\big(F_\theta(h^{(t-1)})\big), \ \ t = 1,\dots,R,
  \label{eq:loop}
\end{equation}
where $\mathrm{Norm}$ is the final normalisation layer, so $R$ steps reach an effective depth of $L \cdot R$ layers with the parameters of $L$ \citep{geiping2025recurrent}.
Ouro ($R=4$, 7.7T pretraining tokens) applies its language-modelling head and next-token loss after every step, so every step ends in a state that the backbone is trained to read.

\subsection{Looped backbone with a readout at every step}
\label{sec:method:loop}
\name{} keeps the decision model of \cref{sec:prelim} and replaces its backbone with a natively looped language model, Ouro or Nanbeige4.2 (\cref{sec:setup}).
\Cref{fig:method}a shows the resulting model, in which the same stack runs $R$ times and one pointer head reads the options after every step.
Applying the pointer head of \cref{eq:pointer} to the hidden states $h^{(t)}$ of every step gives the logits $z^{(t)}$ and one readout $p^{(t)} = \mathrm{softmax}(z^{(t)} / T_t)$ per question and depth, whose answer we write $\hat{o}^{(t)} = \arg\max p^{(t)}$.
The head and the LoRA adapter are shared across steps, and training averages the loss over steps at $T_t = 1$,
\begin{equation}
  \mathcal{L} = \frac{1}{R} \sum_{t=1}^{R} \ell\big(p^{(t)}, y\big),
  \label{eq:steploss}
\end{equation}
so every depth is trained to be a decision model.
Step $t$ depends only on earlier steps, so serving at depth $k$ computes only the first $k$ steps at about $k/R$ of the full cost, and one pass to depth $R$ yields all $R$ readouts, which we use for evaluation.

\label{sec:method:calibration}
Shallow and deep readouts can be miscalibrated by different amounts, so we fit one temperature $T_k$ per depth by minimising the log loss on the calibration partition of the training suite \citep{guo2017calibration}.
No benchmark is used for fitting, and since temperature scaling keeps the predicted option, accuracy at each depth does not depend on $T_k$.

\subsection{Deciding when to go deeper}
\label{sec:method:stopper}
\name{} reads every question at depth one, and after each step a stopper decides which questions still need another one (\cref{fig:method}b).
A further step continues from the hidden states of the last one, so the request is never read again.
A question can only go deeper together with its state, so, as \cref{fig:method}c shows, the state $s$ of a request runs to depth $d(s) = \max_j d(q_j)$.

\paragraph{What the stopper predicts.}
A confidence threshold chosen for accuracy on labelled data transfers poorly for decision models, because shallow readouts are overconfident on new data and blind to the errors that more steps would fix (\cref{sec:analysis:stop}).
At each depth $k < R$, the stopper instead predicts whether the current answer already agrees with the deepest one, the consistency target of CATs,
\begin{equation}
  y_k = \mathbf{1}\big[\hat{o}^{(k)} = \hat{o}^{(R)}\big], \ \ \hat{y}_k = \sigma\big(w_k^{\top} \phi_k + b_k\big),
  \label{eq:stopper}
\end{equation}
a logistic regression per depth.
The features $\phi_k$ come from the readouts at depths $k$ and $k-1$ only: the top probability, the margin between the two most likely options, the normalised entropy, whether the answer equals that at depth $k-1$, $\mathrm{KL}(p^{(k-1)} \,\|\, p^{(k)})$, the change of the top probability, $\log K$, and $k$.
With a threshold $\tau$, question $q$ stops at
\begin{equation}
  d_\tau(q) = \min\big(\{k < R : \hat{y}_k(q) \ge \tau\} \cup \{R\}\big).
  \label{eq:depth}
\end{equation}

\paragraph{Fitting without labels.}
The target $y_k$ is the model's own deepest answer, so fitting the stopper needs no correct answers, and neither does the threshold.
We take the smallest $\tau$, and so the cheapest stops, under which at least 98\% of the final answers $\hat{o}^{(d_\tau(q))}$ agree with the deepest answers $\hat{o}^{(R)}$ on the fitting questions.
A deployment can therefore fit the stopper on the calibration partition together with unlabelled requests from its own traffic, which we emulate with two folds of each test panel split by record group (\cref{app:stopper}).

\begin{table*}[t]
  \centering
  \small
  \setlength{\tabcolsep}{2.9pt}
  \renewcommand{\arraystretch}{1.08}
  \newcolumntype{a}{>{\columncolor{loopblue!7}}c}
  \newcommand{\grouprow}[1]{\multicolumn{8}{>{\columncolor{rowgray}}l}{\textit{#1}} & & \\}
  \begin{tabular}{lccccccc!{\hspace{4pt}}aa}
    \toprule
    \multirow{2}{*}{System} & \multirow{2}{*}{Depth} & Latency\,$\downarrow$ & \multicolumn{3}{c}{Tuning\,$\uparrow$} & \multicolumn{3}{c}{Held-out\,$\uparrow$} & \multicolumn{1}{c}{Benchmark\,$\uparrow$} \\
    \cmidrule(lr){4-6}\cmidrule(lr){7-9}\cmidrule(lr){10-10}
    & & (ms) & \kevseen{} & \kevskills{} & \kevtransfer{} & \kevbreadth{} & \levbench{} & Avg. & JevBench \\
    \midrule
    \grouprow{\name{} (one checkpoint per model, read at every depth)}
    & 1 & \textbf{\phantom{0}15} & 78.7 & 47.6 & 50.9 & 57.0 & 44.5 & 50.8 & 58.0 \\
    & 2 & \phantom{0}30 & 83.7 & 58.2 & 60.4 & 63.4 & 53.0 & 58.2 & 65.4 \\
    & 3 & \phantom{0}44 & 85.7 & 62.5 & 65.9 & 66.4 & 56.4 & 61.4 & 68.0 \\
    \multirow{-4}{*}{\shortstack[l]{\name{}-1.4B\\[-1pt]{\scriptsize\color{black!55}Ouro-1.4B}}} & 4 & \phantom{0}58 & 85.9 & 63.7 & 66.4 & 66.0 & 58.1 & 62.1 & 68.0 \\[2pt]
    & 1 & \phantom{0}51 & 79.0 & 51.9 & 56.6 & 55.7 & 46.9 & 51.3 & 63.2 \\
    & 2 & 101 & 85.7 & 65.9 & 66.1 & 65.3 & 58.6 & 62.0 & 71.0 \\
    & 3 & 150 & \underline{86.2} & 70.4 & 71.3 & 67.1 & \underline{62.6} & \underline{64.8} & \underline{75.3} \\
    \multirow{-4}{*}{\shortstack[l]{\name{}-2.6B\\[-1pt]{\scriptsize\color{black!55}Ouro-2.6B}}} & 4 & 200 & 86.1 & \underline{71.4} & \textbf{72.5} & \textbf{67.7} & \textbf{63.6} & \textbf{65.7} & \textbf{76.2} \\[2pt]
    & 1 & \phantom{0}23 & 82.6 & 59.6 & 62.0 & 62.6 & 51.8 & 57.2 & 69.3 \\
    \multirow{-2}{*}{\shortstack[l]{\name{}-3B\\[-1pt]{\scriptsize\color{black!55}Nanbeige4.2-3B}}} & 2 & \phantom{0}45 & 86.0 & 66.9 & 70.0 & 66.1 & 59.4 & 62.8 & 72.7 \\
    \grouprow{Dense controls, same data and recipe}
    Qwen3-1.7B & -- & \phantom{0}\underline{22} & 81.3 & 50.8 & 58.2 & 58.3 & 48.0 & 53.2 & 63.6 \\
    Qwen3.5-2B & -- & 105 & 83.3 & 63.3 & 62.8 & 64.7 & 53.1 & 58.9 & 68.8 \\
    Qwen3.5-4B & -- & 140 & \textbf{87.9} & \textbf{72.9} & \underline{71.4} & \underline{67.4} & 62.1 & 64.7 & 73.6 \\
    \grouprow{Released references, trained differently}
    {\color{black!55}Kev-0.8B} & {\color{black!55}--} & {\color{black!55}\phantom{0}79} & {\color{black!55}83.4} & {\color{black!55}56.0} & {\color{black!55}57.8} & {\color{black!55}58.9} & {\color{black!55}47.4} & {\color{black!55}53.2} & {\color{black!55}63.6} \\
    {\color{black!55}Kev-4B} & {\color{black!55}--} & {\color{black!55}108} & {\color{black!55}88.1} & {\color{black!55}69.2} & {\color{black!55}73.3} & {\color{black!55}67.7} & {\color{black!55}63.3} & {\color{black!55}65.5} & {\color{black!55}75.8} \\
    {\color{black!55}Jev 1.13.0} & {\color{black!55}--} & {\color{black!55}652\rlap{$^\dagger$}} & {\color{black!55}83.5} & {\color{black!55}77.7} & {\color{black!55}80.7} & {\color{black!55}76.0} & {\color{black!55}69.2} & {\color{black!55}72.6} & {\color{black!55}86.6} \\
    \bottomrule
  \end{tabular}
  \caption{Accuracy (\%) and median latency per request of every system on the six panels of \cref{tab:panels}. Avg.\ is the unweighted mean of \kevbreadth{} and \levbench{}. Among the models trained on identical data, the \name{} models and the dense controls, bold marks the best value and underline the second best. $^\dagger$Measured through the production API with the network included, and not comparable to the local measurements.}
  \label{tab:main}
\end{table*}

\paragraph{Cost.}
The cost of a request is the depth $d(s)$ of its state divided by $R$, and we average it over requests weighted by their length in tokens.
A fixed depth $k$ costs $k/R$, and serving a share of the requests at depth $k$ and the rest at depth $k+1$ reaches any cost in between, with the accuracies mixed in the same shares.
We call this piecewise-linear curve the \emph{fixed-depth frontier}, and on-demand depth is useful only if it is more accurate than the frontier at the same cost.

%% file: sections/setup.tex
\section{Experimental Setup}
\label{sec:setup}

\paragraph{Models.}
We build \name{} on three looped backbones.
Ouro-1.4B and Ouro-2.6B run their stack $R=4$ times and were pretrained with a loss after every step.
Nanbeige4.2-3B runs its stack twice and was pretrained with a loss on its last step only, so it tests whether \name{} needs intermediate steps that were supervised in pretraining.
We call the three models \name{}-1.4B, \name{}-2.6B, and \name{}-3B.
The dense controls are Qwen3-1.7B-Base \citep{yang2025qwen3}, Qwen3.5-2B-Base, and Qwen3.5-4B-Base.
As references we report the released Kev-0.8B and Kev-4B and the closed Jev 1.13.0.

\paragraph{Training.}
Every model is trained with LoRA \citep{hu2022lora} of rank 16 for one epoch on the same 31,329 records, which merge the training sources of Kev-0.8B into a single stage.
No label comes from Jev, so \name{} is not distilled from the model it is compared with.
\Cref{app:setup} gives the settings and the data.
Training uses up to four H200 GPUs per model.

\paragraph{Evaluation panels.}
We fixed the recipe and chose the design of the stopper on three tuning panels from frozen Kev suites, \kevseen{} (training sources), \kevtransfer{} (new sources), and \kevskills{} (held-out skill templates).
Our models read the two held-out panels only after that.
\levbench{} is a suite we build and froze before any model read it. Its tiers move step by step away from the training data, from new items of the training sources to datasets never trained on, and \cref{app:levbench} describes its construction.
\kevbreadth{} holds 14 datasets from five application areas that Kev never trains on.
We also report the 231 public JevBench items, of which 48 are easy, 72 standard, and 111 hard.
\Cref{tab:panels} lists all panels.

\paragraph{Metrics.}
We report accuracy on every panel and the expected calibration error (ECE) on the hard tier of JevBench, which we score with JevBench's own code (\cref{app:evaluator}).
For on-demand depth we also report the cost of \cref{sec:method:stopper}.
Latency is the median per request on one H200 GPU in bf16.
Confidence intervals are 95\% paired bootstrap intervals that resample groups of related questions.

%% file: sections/results.tex
\newpage
\section{Main Results}
\label{sec:results}

\begin{wrapfigure}[17]{r}{0.36\textwidth}
  \centering
  \vspace{-12pt}
  \includegraphics[width=0.36\textwidth]{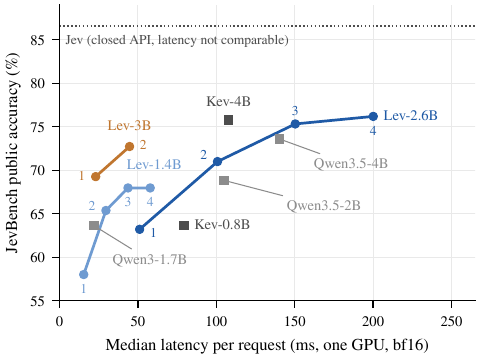}
  \caption{Accuracy and median latency on the 231 public JevBench items. Each \name{} checkpoint gives one operating point per depth, labelled by depth. Squares are separate dense checkpoints.}
  \label{fig:pareto}
\end{wrapfigure}
\Cref{tab:main} supports four conclusions.
(1)~Extra loop steps raise accuracy on the same weights, with diminishing returns.
The gain holds for every backbone and on every panel and reaches 13.0 points on JevBench for \name{}-2.6B (\cref{fig:paired}), while the fourth step of the Ouro models adds at most one point.
Most of the gain comes with the first extra step, which lifts the held-out average of \name{}-2.6B from 51.3\% to 62.0\%, against 65.7\% at depth four.
It is also smallest on the training sources, as \name{}-1.4B and \name{}-2.6B gain about 7 points on \kevseen{} from depth one to depth four, against 9.0 to 19.5 points on the four panels further from the training data.
(2)~One checkpoint covers several operating points and can beat a dense control at the same latency.
\name{}-3B at depth one is 4.0 points above Qwen3-1.7B on the held-out average at 23 against 22\,ms, and \name{}-2.6B at depth two is 3.1 points above Qwen3.5-2B at 101 against 105\,ms, while Qwen3-1.7B is level with or above \name{}-1.4B (\cref{fig:pareto}).
(3)~\name{} matches larger models with fewer parameters but not fewer operations.
\name{}-3B is level with Qwen3.5-4B on JevBench, and \name{}-2.6B with the released Kev-4B (\cref{fig:paired}), but each loop step runs the whole backbone.
(4)~On-demand depth saves computation at almost no cost in accuracy and beats dense cascades at low latency (\cref{sec:analysis:ondemand,sec:analysis:cascade}).

%% file: sections/analysis.tex
\section{Analysis}
\label{sec:analysis}

We analyse \name{} through five research questions.
RQ1 asks where extra depth helps, by difficulty tier, by revision between steps, and by application area.
RQ2 asks whether on-demand depth is more accurate than a fixed mix of depths at the same cost, and RQ3 asks which stopping rule makes it so.
RQ4 compares \name{} with dense models and cascades at the same measured latency, and RQ5 asks how joint training, more data, and full-weight fine-tuning change the accuracy at each depth.

\subsection{RQ1: Where does extra depth help?}
\label{sec:analysis:depth}

\begin{figure*}[!b]
  \centering
  \includegraphics[width=\textwidth]{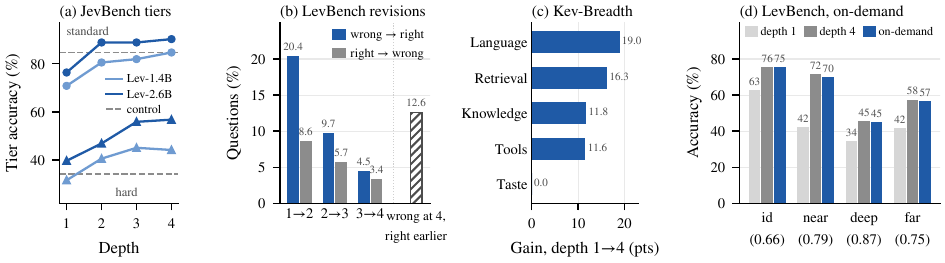}
  \caption{Where depth helps. (a)~JevBench accuracy on the standard and hard tiers, with the Qwen3-1.7B control dashed. (b)~\levbench{} questions that \name{}-2.6B revises between consecutive depths, or gets right at some shallower depth but wrong at depth four. (c)~Gain of \name{}-2.6B from depth one to four per \kevbreadth{} area. (d)~\name{}-2.6B on the \levbench{} tiers, with the cost of on-demand depth in parentheses.}
  \label{fig:where}
\end{figure*}

\paragraph{Gains concentrate on hard items.}
From depth one to depth four, \name{}-2.6B gains a net 10 of the 72 standard JevBench items and 19 of the 111 hard items, while the easy tier is nearly solved at every depth (\cref{fig:where}a).
Most revisions are corrections, as 29 hard items become correct and 10 wrong.

\paragraph{Depth revises answers, and helps some areas more than others.}
On \levbench{}, the first extra step of \name{}-2.6B fixes 20.4\% of the questions and breaks 8.6\% (\cref{fig:where}b), and later steps change fewer.
For 12.6\% of the questions some shallower depth is right while the deepest is wrong, and a stopper that predicts agreement with the deepest answer cannot recover them.
\Cref{fig:where}c shows that on \kevbreadth{} depth adds 19.0 points on language understanding but nothing on the two taste tasks, Humicroedit and cfcolor.
Extra steps likely add the serial computation that combining facts needs, which taste does not use.

\subsection{RQ2: Does on-demand depth beat a fixed depth?}
\label{sec:analysis:ondemand}

\begin{wrapfigure}[19]{r}{0.45\textwidth}
  \centering
  \includegraphics[width=0.45\textwidth]{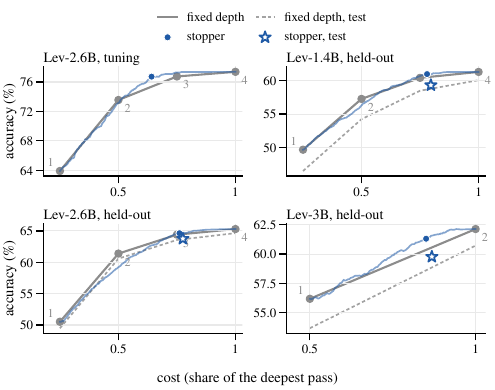}
  \caption{Accuracy against cost. Gray points and line are the fixed depths and their frontier, and the blue line moves the threshold of the stopper to the chosen dot. The dashed line and the star repeat both on the locked test partitions.}
  \label{fig:adaptive}
\end{wrapfigure}

\paragraph{On-demand depth computes less at almost no accuracy cost.}
\Cref{fig:adaptive} shows that on the held-out panels the stopper stays within one point of the deepest readout while computing 76\% to 85\% of its cost, and \cref{tab:ondemand} gives the pooled numbers.
The same holds on every panel of \cref{tab:adaptive} except the 231 JevBench items.

\begin{table}[!t]
  \centering
  \small
  \renewcommand{\arraystretch}{1.08}
  \setlength{\tabcolsep}{4.5pt}
  \begin{tabular}{lccccc}
    \toprule
    Model & Deepest & On-demand & Cost & vs.\ deepest & vs.\ frontier \\
    \midrule
    \rowcolor{rowgray} \multicolumn{6}{l}{\textit{Development}} \\
    \name{}-1.4B & 61.3 & 61.0 & 0.78 & $-0.3$ \ci{$[-0.6, -0.0]$} & $+0.5$ \ci{$[-0.0, +1.0]$} \\
    \name{}-2.6B & 65.3 & 64.6 & 0.76 & $-0.7$ \ci{$[-1.0, -0.4]$} & $+0.2$ \ci{$[-0.4, +0.8]$} \\
    \name{}-3B & 62.1 & 61.3 & 0.85 & $-0.8$ \ci{$[-1.1, -0.5]$} & $+1.0$ \ci{$[+0.6, +1.3]$} \\
    \rowcolor{rowgray} \multicolumn{6}{l}{\textit{Locked test}} \\
    \name{}-1.4B & 60.1 & 59.3 & 0.80 & $-0.7$ \ci{$[-1.0, -0.5]$} & $+0.5$ \ci{$[+0.1, +0.9]$} \\
    \name{}-2.6B & 64.6 & 63.7 & 0.78 & $-0.9$ \ci{$[-1.1, -0.6]$} & $+0.1$ \ci{$[-0.4, +0.5]$} \\
    \name{}-3B & 60.7 & 59.8 & 0.87 & $-1.0$ \ci{$[-1.2, -0.7]$} & $+0.9$ \ci{$[+0.6, +1.2]$} \\
    \bottomrule
  \end{tabular}
  \caption{On-demand depth on the two held-out panels, pooled over their questions. Cost is the token-weighted share of the deepest pass. The last two columns are paired differences in accuracy points with 95\% CIs. \Cref{tab:adaptive,tab:test} give every panel.}
  \label{tab:ondemand}
\end{table}

\paragraph{The gain over the fixed-depth frontier depends on its shape.}
Against a fixed depth of the same cost, the gain is small.
On-demand \name{}-2.6B is 1.4 points above the fixed-depth frontier on the tuning panels but only 0.2 points on the held-out panels, while \name{}-3B gains 1.0 point there.
In \cref{fig:adaptive} the frontier of the Ouro models is concave because the first extra step brings most of the gain, whereas the two-step frontier of \name{}-3B is a straight line that most stopping signals beat.

\paragraph{The locked test confirms the development result.}
On the test partitions of both held-out panels, read once after the backbones, the recipe, and the stopper were fixed, on-demand depth again loses at most 1.0 point against the deepest readout, is above the frontier for \name{}-1.4B and \name{}-3B, and is level with it for \name{}-2.6B (stars in \cref{fig:adaptive}).

\begin{figure}[!t]
  \centering
  \includegraphics[trim=0 0 147bp 16.8bp, clip, width=0.72\textwidth]{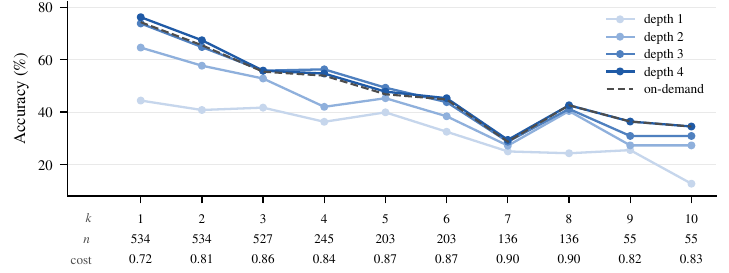}
  \caption{\name{}-2.6B on \levbench{} by difficulty $k$, such as the number of reasoning steps, on the generated reasoning families and the public datasets that define one, with the number of questions $n$ and the cost of on-demand depth below each $k$. Values of $k$ above 10 cover 22 questions and are left out. Cost is the token-weighted share of the deepest pass.}
  \label{fig:difficulty}
\end{figure}

\paragraph{Harder and less familiar items get more steps.}
\label{sec:analysis:tiers}
The stopper follows difficulty without seeing it.
On JevBench, on-demand \name{}-2.6B uses 1.2, 2.4, and 3.0 steps on easy, standard, and hard items (\cref{fig:teaser}a).
On \levbench{} it computes 66\% of the deepest pass on the \emph{id} tier, 1.8 points above the frontier, and 75\% to 87\% on the tiers further from the training data (\cref{fig:where}d), and the same holds on further panels and on harder generated questions (\cref{tab:extra,fig:difficulty}).

\subsection{RQ3: How should the depth be chosen?}
\label{sec:analysis:stop}

\begin{wrapfigure}[18]{r}{0.45\textwidth}
  \centering
  \vspace{-12pt}
  \includegraphics[width=0.45\textwidth]{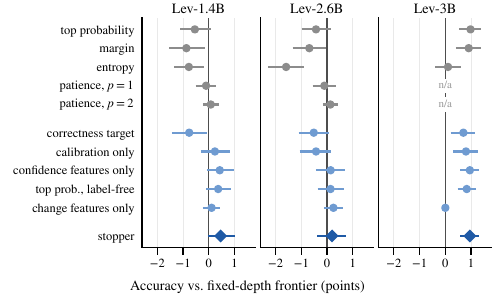}
  \caption{Accuracy of each stopping rule minus the fixed-depth frontier at the same cost on the pooled held-out panels, with 95\% CIs. Gray rows are early-exit rules, and patience needs at least three steps. Light blue rows change one part of the stopper or set a single top-probability threshold the way the stopper's is set, and the change features alone never stop \name{}-3B.}
  \label{fig:ablation}
\end{wrapfigure}

\paragraph{Confidence thresholds set on labelled data do not beat the frontier.}
The usual early-exit rule stops once the top probability reaches a threshold chosen for accuracy on the labelled calibration partition.
For the Ouro models, \cref{fig:ablation} shows that neither this rule nor other early-exit rules, thresholds on the margin or entropy and patience-based exit \citep{zhou2020pabee}, beat the frontier.

\paragraph{Shallow confidence is overconfident and blind to fixable errors.}
Two properties of the shallow readouts explain this (\cref{tab:diagnostic}).
First, they are overconfident on new data.
The questions that \name{}-2.6B stops at depth one are 94.8\% correct on the calibration partition but only 78.9\% on \kevtransfer{}, at almost the same confidence.
Second, confidence separates right from wrong answers with an AUROC of 0.67 to 0.83, but separates the wrong answers that more steps fix from the rest with only 0.46 to 0.61.

\begin{figure}[!t]
  \centering
  \includegraphics[width=\textwidth]{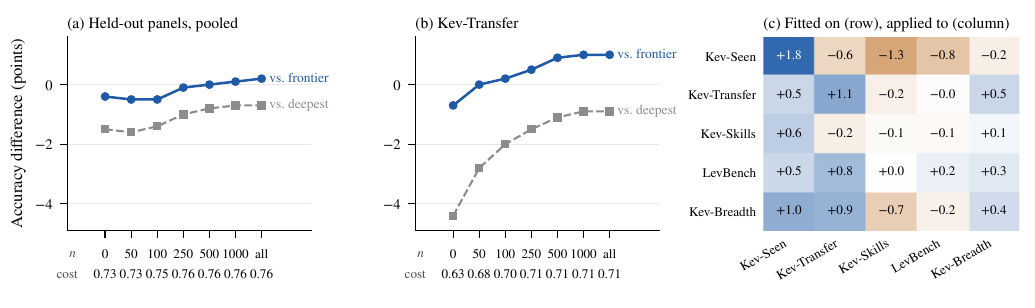}
  \caption{The stopper of \name{}-2.6B with few unlabelled requests. (a, b) Fitted with $n$ unlabelled questions per fold, as the accuracy difference in points to the deepest readout and to the fixed-depth frontier, averaged over five draws, with the cost of on-demand depth below each $n$. (c) Fitted on the unlabelled requests of one panel and applied to another, as the difference to the frontier. On the diagonal, it is fitted and applied on the same panel.}
  \label{fig:unlabelled}
\end{figure}

\paragraph{Setting the threshold on unlabelled requests carries the stopper.}
The stopper avoids the first problem by setting its threshold for agreement with the deepest answer on unlabelled requests from the panel it serves.
No signal we tried overcomes the second problem, which limits how far any stopper can beat the frontier.
\Cref{fig:ablation} shows that this carries its gain.
Set the same way, a single top-probability threshold comes within 0.2 points of the stopper, so the features matter little, while a correctness target fitted on the labelled calibration partition falls below the frontier for both Ouro models.
A few hundred unlabelled questions suffice, as 500 per fold bring \name{}-2.6B within 0.2 points of the full fit (\cref{app:unlabelled}).
\Cref{fig:unlabelled} shows this, and a stopper fitted on one panel stays within 0.7 points of the frontier on the others unless it is fitted on the training sources.

\subsection{RQ4: Does \name{} beat dense cascades at the same latency?}
\label{sec:analysis:cascade}

\begin{figure}[t]
  \centering
  \begin{minipage}[t]{0.485\textwidth}
    \centering
    \includegraphics[width=\linewidth]{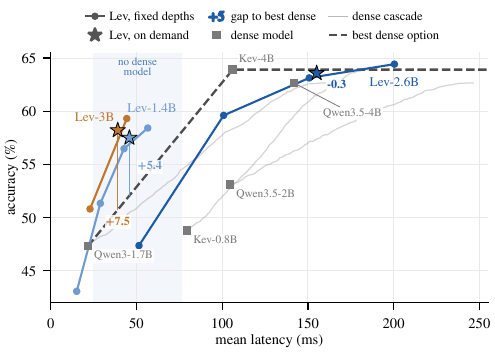}
    \caption{Accuracy against mean latency on the \levbench{} test partition. Lines join the fixed depths of one \name{} checkpoint, and stars mark its on-demand depth. The dashed line is the best dense model or cascade, and each number is the gap of on-demand \name{} to it.}
    \label{fig:cascade}
  \end{minipage}\hfill
  \begin{minipage}[t]{0.485\textwidth}
    \centering
    \includegraphics[width=\linewidth]{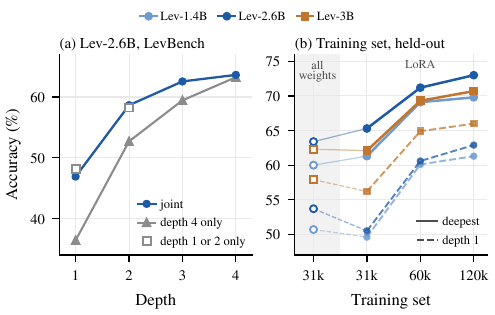}
    \caption{Training choices. (a)~\name{}-2.6B on \levbench{}, trained jointly, at depth four only, or for depth one or two alone. (b)~Pooled held-out accuracy by training set. The shaded column trains all weights on \trainsmall{}.}
    \label{fig:training}
  \end{minipage}
\end{figure}
\paragraph{Small \name{} models fill the low-latency range.}
Latency is measured on the locked test partitions, and an escalated cascade request pays for both of its models (\cref{app:cascade}).
\Cref{fig:cascade} shows that the small \name{} models open a range of latencies that no dense option covers well.
At 39 to 46\,ms, on-demand \name{}-1.4B and \name{}-3B are 7.0 to 8.8 points more accurate than the only cascade that reaches this latency, and 5.4 to 7.5 points above the best dense option.
\name{}-2.6B at about 150\,ms is no better than Kev-4B alone.

\paragraph{The advantage is in latency, not computation.}
Part of this advantage comes from a gap in the dense options, none of which answers between 22 and 79\,ms.
Counted in floating-point operations on the development partitions, every \name{} model is at or below the dense options, because each loop step runs the whole backbone (\cref{app:cascade}).

\subsection{RQ5: How do training choices shape the depth curve?}
\label{sec:analysis:scale}

\paragraph{Joint training serves every depth at little cost.}
\label{sec:analysis:joint}
\Cref{fig:training}a compares the step-averaged loss of \cref{eq:steploss} with Ouro-2.6B models trained for a single depth, and \cref{tab:joint} gives the paired differences.
The jointly trained \name{}-2.6B is level with them at depths two and four and at most 1.3 points behind at depth one, while training only the last step leaves the shallow readouts unusable (\cref{app:joint}).

\paragraph{More data mainly strengthens the shallow readouts.}
All other results use 31,329 training records (\trainsmall{}), which we extend to 60k and 120k (\trainmid{} and \trainlarge{}, \cref{app:trainsets}).
On the pooled held-out panels, depth one of \name{}-2.6B rises from 50.5\% to 62.9\% and the deepest readout from 65.3\% to 73.0\%, so on-demand cost falls from 0.76 to 0.63 (\cref{fig:training}b).
Part of this rise is in distribution, because the larger sets add sources that \levbench{} also draws on.
On \kevbreadth{}, which no training set touches, the deepest readout stays between 67.3\% and 67.7\%, while Qwen3.5-2B gains 0.4 points and Qwen3.5-4B 2.0 points (\cref{tab:dense-scale}).

\paragraph{LoRA beats full-weight fine-tuning on the Ouro models.}
Training all weights, shaded in \cref{fig:training}b, lowers the deepest readout of both Ouro models by 1.3 to 1.9 points and removes their gain over the frontier, while \name{}-3B is unchanged (\cref{tab:scale}).
A Brier term, step weights that grow with the step, and more loop steps than in pretraining change the deepest readout by at most 0.9 points and raise its gain over the frontier by at most 0.1 points (\cref{tab:recipe}).
\label{sec:analysis:calibration}
The hard-tier ECE of every readout lies between 0.18 and 0.26 after temperature scaling with no trend in depth, as for the Qwen3-1.7B control, so the gap to Kev and Jev comes from our single-stage recipe rather than from the loop (\cref{tab:calibration}).

%% file: sections/conclusion.tex
\section{Conclusion}
\label{sec:conclusion}

We build \name{}, a decision model that thinks in loop steps rather than tokens.
One checkpoint serves several operating points whose accuracy rises with depth, mostly on hard questions.
Setting the stopping threshold on unlabelled requests saves 15\% to 24\% of the computation at a cost of under one accuracy point, and below 50\,ms the 1.4B and 3B models are 5.4 to 8.8 points more accurate than dense models and cascades of the same latency.
The stopper spends more steps on harder questions without seeing their difficulty, and predicting agreement with the deepest answer works where confidence thresholds set on labelled data do not.
These results suggest that a natively looped backbone turns depth into a setting chosen at serving time, so one decision model can trade latency for accuracy per question instead of per deployment.
Two directions follow from the analysis.
A stopping target that also recovers the questions a shallower step answers correctly could add accuracy beyond the deepest readout, and larger looped backbones would show whether the gains on hard questions grow with scale.

%% file: sections/limitations.tex
\section*{Limitations}

Our study covers looped backbones of 1.4B to 3B parameters, the sizes that are publicly available, and applying the same recipe to larger looped models would show how far the depth curve extends.
We report JevBench results on its 231 public items, and further decision benchmarks would broaden the picture.
We count the cost of on-demand depth in loop steps, and a serving path that batches requests at different depths would turn this saving into wall-clock time.
Finally, latency is measured one request at a time without fused kernels, graph capture or state caching for all systems, and optimised serving for looped and dense models alike is left to future work.

%% file: sections/appendix.tex
\section{Motivation Details}
\label{app:motivation}

\subsection{Routing from Kev-0.8B to Kev-4B}
\begin{figure}[H]
  \centering
  \includegraphics[width=\colfig]{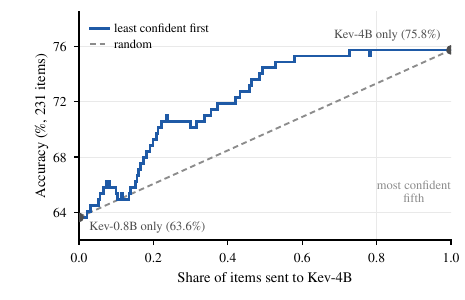}
  \caption{Accuracy on the 231 public JevBench items when the items on which Kev-0.8B is least confident are answered by Kev-4B, against random routing.}
  \label{fig:routing}
\end{figure}

\Cref{fig:routing} is the full routing curve behind \cref{fig:motivation}b.
Items go from Kev-0.8B to Kev-4B in order of increasing top probability of Kev-0.8B, and the curve shows JevBench accuracy at every share of items sent.
Confidence routing is above random routing once more than a seventh of the items are sent.
The median top probability of Kev-0.8B is 0.80 on the items that both models answer correctly and 0.44 on the items that only Kev-4B answers correctly.
The curve is flat over the last fifth, because Kev-4B changes no outcome among the fifth of items where Kev-0.8B is most confident.

\subsection{Generated Reasoning}
\Cref{fig:motivation}c takes the hard accuracy and the median latency of the hard items from the results that JevBench publishes for djev and OpenJev.
Each server is measured on its own endpoint, so the latencies compare the two modes of one server but not the two servers.
Both servers run DiffusionGemma 26B. djev itself always answers with a single read, so its thinking mode is a generation path over the same bf16 weights that JevBench added, while OpenJev has its own thinking option and uses NVFP4-quantised weights for the single read and the bf16 weights for thinking.
The token counts are the generated tokens per decision that JevBench reports for each thinking mode.
Thinking also fails in ways a single read does not. The thinking mode of djev gets 2 of the 48 easy items wrong, and with a budget of 8,192 generated tokens it returned no valid answer for 72 of the 534 decisions of the full benchmark, which also contains non-public items.

\subsection{Retrofitted Loops}
\label{app:retrofit}
\Cref{fig:motivation}d summarises \cref{tab:retrofit}, which loops layers of Qwen3.5-0.8B-Base, the backbone of Kev-0.8B.
In the default loop, layers 8 to 15 of the 24-layer stack run twice.
Every pass after the first is gated per channel in the ReZero style \citep{bachlechner2021rezero},
\begin{equation}
  h \leftarrow h + g_{p,i} \odot \big(f_i(h) - h\big),
\end{equation}
where $f_i$ is layer $i$ and the gate $g_{p,i}$ starts at zero.
At initialisation the looped model is therefore exactly the pretrained model, and training decides how much of the extra depth to use.
We train a LoRA adapter and a pointer head over the options (\cref{sec:prelim}) with Kev's recipe on the 12,576 records of Kev's main training partition.
The recipe is two epochs of LoRA with rank 16 and a peak learning rate of $10^{-4}$.
The adapter is shared across passes, and each pass keeps its own attention and recurrent caches.
We read the development partitions of the three tuning panels (\cref{tab:panels}).
No variant in \cref{tab:retrofit} improves accuracy on any panel with a confidence interval above zero.
After training, the gates of the default loop have a mean magnitude of about 0.006, so the second pass contributes less than one percent of its residual update.
A loop that starts closed receives little gradient to open, and a loop that starts open moves hidden states away from what the later layers were pretrained on, so removing the gate lowers accuracy.

\begin{table}[!htbp]
  \centering
  \small
  \renewcommand{\arraystretch}{1.08}
  \begin{tabular}{lcccc}
    \toprule
    Variant & Seeds & \kevseen{} & \kevtransfer{} & \kevskills{} \\
    \midrule
    Baseline accuracy & 3 & 81.1 & 56.4 & 34.1 \\
    \midrule
    \rowcolor{loopblue!6} Default loop & 3 & $-0.08$ \ci{$[-0.55, +0.40]$} & $+0.48$ \ci{$[-0.29, +1.43]$} & $+0.18$ \ci{$[-0.64, +0.93]$} \\
    3 passes & 1 & $-0.47$ \ci{$[-1.27, +0.40]$} & $-0.29$ \ci{$[-1.43, +0.86]$} & $-0.83$ \ci{$[-2.14, +0.55]$} \\
    16 layers & 1 & $-0.32$ \ci{$[-1.34, +0.63]$} & $+0.19$ \ci{$[-1.24, +1.53]$} & $-1.20$ \ci{$[-2.52, +0.09]$} \\
    Gate lr $\times 3$ & 1 & $+0.00$ \ci{$[-0.87, +0.87]$} & $-0.57$ \ci{$[-1.72, +0.57]$} & $-1.11$ \ci{$[-2.18, -0.09]$} \\
    Gate lr $\times 10$ & 1 & $-0.47$ \ci{$[-1.42, +0.40]$} & $-0.10$ \ci{$[-1.34, +1.24]$} & $-0.92$ \ci{$[-2.13, +0.28]$} \\
    Gate lr $\times 30$ & 1 & $+0.71$ \ci{$[-0.16, +1.66]$} & $-0.10$ \ci{$[-1.43, +1.24]$} & $-1.02$ \ci{$[-2.30, +0.19]$} \\
    \rowcolor{loopblue!6} No gate & 1 & $-0.55$ \ci{$[-1.82, +0.71]$} & $-2.20$ \ci{$[-4.02, -0.48]$} & $-0.92$ \ci{$[-2.80, +1.00]$} \\
    \bottomrule
  \end{tabular}
  \caption{Retrofitted loops on Qwen3.5-0.8B-Base. The first row is the accuracy (\%) of the unlooped baseline, and every other cell is the paired accuracy difference to it in points, with a 95\% CI, on development partitions. The default loop runs layers 8--15 twice behind gates trained with a learning rate of $10^{-3}$. 3 passes runs layers 8--15 three times, 16 layers loops layers 4--19, gate lr $\times N$ multiplies the gate learning rate by $N$, and no gate removes the gates.}
  \label{tab:retrofit}
\end{table}

\section{Setup Details}
\label{app:setup}

\subsection{Training Settings}
\label{app:hparams}

\Cref{tab:hparams} lists the training settings.
The delimiters of \cref{sec:prelim} are special tokens of the backbone's tokenizer, and user text that spells them is rewritten so that it cannot forge a boundary.

\begin{table}[!htbp]
  \centering
  \small
  \renewcommand{\arraystretch}{1.08}
  \setlength{\tabcolsep}{5pt}
  \begin{tabular}{ll}
    \toprule
    Setting & Value \\
    \midrule
    \rowcolor{rowgray} \multicolumn{2}{l}{\textit{Adapter and head}} \\
    LoRA rank / $\alpha$ / dropout & 16 / 32 / 0.05 \\
    LoRA targets & attention + MLP \\
    Pointer head dimension $d_p$ & 256 \\
    \rowcolor{rowgray} \multicolumn{2}{l}{\textit{Optimisation}} \\
    Optimiser & AdamW, weight decay 0.01 \\
    Schedule & one-cycle, 10\% warm-up \\
    Peak learning rate & $10^{-4}$ \\
    Records per step & 64 \\
    Micro-batch size & 8 on average \\
    Epochs & 1 \\
    Precision & bf16 autocast \\
    Step weights $w_t$ (looped) & uniform, $1/R$ \\
    Temperature grid $T_k$ & 121 log-spaced in $[0.25, 4]$ \\
    \rowcolor{rowgray} \multicolumn{2}{l}{\textit{Data}} \\
    Maximum state length & 5,120 tokens \\
    Records, \trainsmall{} & 31,329 \\
    Records, \trainmid{} & 60k \\
    Records, \trainlarge{} & 120k \\
    \rowcolor{rowgray} \multicolumn{2}{l}{\textit{Full-weight variant}} \\
    Backbone / head lr & $10^{-5}$ / $10^{-4}$ \\
    Weights / optimiser & bf16 / fp32 \\
    \bottomrule
  \end{tabular}
  \caption{Training settings of \name{} and the dense controls. The step weights apply only to \name{}, and the last block only to the full-weight runs.}
  \label{tab:hparams}
\end{table}

\paragraph{Backbones.}
Ouro-1.4B is a 24-layer decoder with hidden size 2048, Ouro-2.6B has 48 layers, and Nanbeige4.2-3B runs 22 layers twice.
The Qwen3-1.7B control lies between the two Ouro backbones in size, so it is the dense model of comparable size for both.

\paragraph{Training.}
Micro-batches group records of similar length so that short records share a pass and long records run alone.
Each optimiser step still sees 64 records.
With probability 0.25, Kev's ``none of the above'' augmentation adds such an option to a choice question and trains two copies of the record, one with the correct option and one without it, where ``none of the above'' is correct.

\subsection{Evaluation Panels}
\label{app:data}

\Cref{tab:panels} maps each panel to its source suite.
The three tuning panels and \kevbreadth{} are development partitions of frozen Kev suites, and \levbench{} is the suite we built for this study.
The calibration partition is not a panel.
It fits the temperature of each depth and the thresholds of the early-exit rules, and it is part of the data the stopper is fitted on.
The locked test uses the test partitions of \levbench{} (8,740 scored questions) and \kevbreadth{} (3,031), and also reports the test partitions of the suites behind the three tuning panels.

\begin{table}[!htbp]
  \centering
  \small
  \renewcommand{\arraystretch}{1.08}
  \setlength{\tabcolsep}{4pt}
  \begin{tabular}{lllrp{7.2cm}}
    \toprule
    Group & Panel & Suite, partition & Questions & Content \\
    \midrule
    \multirow{3}{*}{Tuning} & \kevseen{} & \texttt{decision-v7}, dev. & 1,464 & Ten public classification and inference datasets, generated policies and rules \\
    & \kevtransfer{} & \texttt{transfer-v9}, dev. & 1,264 & New sources such as MMLU-Pro and PAWS, held-out rule families, unknowable items \\
    & \kevskills{} & \texttt{hard-v1}, dev. & 1,083 & Seven programmatically labelled skill families, held-out templates \\
    \midrule
    \multirow{2}{*}{Held-out} & \levbench{} & ours, dev. & 4,370 & Tiers \emph{id}, \emph{near}, \emph{deep}, and \emph{far} by distance from the training data \\
    & \kevbreadth{} & \texttt{breadth-v1}, dev. & 3,023 & 14 datasets from five application areas, never trained on \\
    \midrule
    Benchmark & JevBench & public items & 231 & 48 easy, 72 standard, and 111 hard items \\
    \midrule
    \color{black!55}Calibration & \color{black!55}-- & \color{black!55}\texttt{decision-v7}, calib. & \color{black!55}1,148 & \color{black!55}Temperatures per depth, and part of the data the stopper is fitted on \\
    \bottomrule
  \end{tabular}
  \caption{Evaluation panels and the suites they come from. Suite names are those of the Kev repository. Questions are the scored questions of each panel.}
  \label{tab:panels}
\end{table}

\subsection{Training Sets}
\label{app:trainsets}

\paragraph{Base training set.}
\trainsmall{} combines Kev's main training partition (12,576 records from public classification and inference datasets, generated policies, and generated rule structures), generated records that state day counts or omit the deciding evidence (2,225), Kev's generated skill families (6,000), developer-tooling decisions (5,320), and consumer-complaint documents (5,219).
After removing records whose state exceeds 5,120 tokens, 31,329 records remain.

\paragraph{Larger training sets.}
\trainmid{} and \trainlarge{} contain \trainsmall{} and add records from a pool that we screened against every evaluation partition we use, including the public JevBench items.
The pool holds 32,000 records from four generated reasoning families (rule application, dependency chains, ledger steps, and date arithmetic), each with a difficulty $k$ inside its training range and 10\% of them unknowable, 12,000 records from new seeds of the skill generators, 16,000 further records from the public sources of \kevseen{}, 7,721 records from ARC, OpenBookQA, and CommonsenseQA, and 22,002 records from CondaQA, Social IQa, CosmosQA, MathQA, QASC, and PIQA.
Duplicates and records that match an evaluation item are removed.
We interleave the pool by part so that every prefix keeps the proportions of the parts.
\trainmid{} and \trainlarge{} add its first 28,660 and 88,660 records, so each larger set contains the smaller one with the same mix (\cref{tab:trainsets}).

\begin{table}[H]
  \centering
  \small
  \renewcommand{\arraystretch}{1.08}
  \setlength{\tabcolsep}{4pt}
  \begin{tabular}{>{\raggedright\arraybackslash}p{3.7cm}rr}
    \toprule
    Part of the pool & \trainmid{} & \trainlarge{} \\
    \midrule
    Generated reasoning families & 10,222 & 31,621 \\
    Six further public datasets & 7,028 & 21,741 \\
    Public sources of \kevseen{} & 5,111 & 15,810 \\
    New seeds, skill generators & 3,833 & 11,858 \\
    ARC, OpenBookQA, CommonsenseQA & 2,466 & 7,630 \\
    \midrule
    All & 28,660 & 88,660 \\
    \bottomrule
  \end{tabular}
  \caption{Records that the larger training sets add to \trainsmall{}, by part of the pool.}
  \label{tab:trainsets}
\end{table}

\subsection{\levbench{}}
\label{app:levbench}

The development partition has 4,870 questions in five tiers, of which 4,370 are scored for accuracy.
The \emph{id} tier (1,370 questions) holds new items of the sources of \trainlarge{}: 970 from its public datasets, 520 of them from sources that \trainsmall{} also trains on, and 400 from the four generated families with a difficulty inside the training range.
The \emph{near} and \emph{deep} tiers (800 questions each) hold a held-out template of the four generated families above, with a difficulty inside and beyond the range that \trainmid{} and \trainlarge{} train on.
\trainsmall{} trains on none of these families, so for \trainsmall{} they are new tasks.
The \emph{far} tier holds eight datasets that never enter training: BoardgameQA, StepGame, CLadder, LogiQA 2.0, AR-LSAT, QuAIL, TimeDial, and StrategyQA.
We score 1,400 of its questions and leave out its 200 StrategyQA questions.
The unknowable tier (300 questions) removes the deciding fact from generated questions and is scored only by how often the model answers confidently.
Every item was screened against \trainsmall{}, all evaluation partitions of the Kev suites, the public JevBench items, and MMLU-Pro.
A public item was removed on any match, and a generated item on an exact match.
The suite was frozen before any model read it.

\subsection{Evaluation Protocol}
\label{app:evaluator}

\paragraph{Evaluator.}
Our evaluator builds each item's request with JevBench's own question builder, passes it through the same request conversion that a Kev server uses, and scores the answer with JevBench's scoring code.
\Cref{tab:evaluator} compares it with the official JevBench harness on the released Kev checkpoints.
On Kev-4B it reproduces the harness exactly, and on Kev-0.8B it matches overall accuracy and differs by under one point on the hard tier.
For a looped model it reads every depth from one forward pass.

\begin{table}[H]
  \centering
  \small
  \renewcommand{\arraystretch}{1.08}
  \begin{tabular}{lcccc}
    \toprule
    & \multicolumn{2}{c}{Kev-0.8B} & \multicolumn{2}{c}{Kev-4B} \\
    \cmidrule(lr){2-3}\cmidrule(lr){4-5}
    & Ours & Harness & Ours & Harness \\
    \midrule
    Acc.\ (\%) & 63.6 & 63.6 & 75.8 & 75.8 \\
    Hard acc.\ (\%) & 36.9 & 36.0 & 54.1 & 54.1 \\
    Hard ECE & 0.172 & 0.181 & 0.112 & 0.112 \\
    \bottomrule
  \end{tabular}
  \caption{Our in-process evaluator against the official JevBench harness on the released Kev checkpoints.}
  \label{tab:evaluator}
\end{table}

\paragraph{Metrics.}
The Intelligence score of \cref{tab:jevbench} follows JevBench version 1.3. It is a weighted mean over tiers of the chance-corrected accuracy $(a-c)/(1-c)$, where $a$ is the accuracy on a tier and $c$ the accuracy of random guessing on it, with the official tier weights renormalised over the three public tiers.
The calibration error is the top-label ECE.
Accuracy is computed in fp32.
On JevBench we add an exact McNemar test to the paired bootstrap.

\paragraph{Latency.}
Latency in \cref{tab:main,tab:jevbench} is the median over the 231 JevBench items of one request at a time, after a short warm-up.

\paragraph{Jev.}
Jev's per-item outcomes are taken from the outcomes published in the JevBench repository, and on \kevskills{} and \kevbreadth{} from the reads published in the Kev repository.

\section{JevBench Results}
\label{app:jevbench}

\subsection{Tiers and Latency}

\Cref{tab:jevbench} splits JevBench by tier (\cref{sec:analysis:depth}).

\begin{table}[!htbp]
  \centering
  \small
  \renewcommand{\arraystretch}{1.08}
  \setlength{\tabcolsep}{5pt}
  \begin{tabular}{lcccccccc}
    \toprule
    System & Params & Depth & Latency (ms) & Easy & Standard & Hard & All & Intelligence \\
    \midrule
    \rowcolor{rowgray} \multicolumn{9}{l}{\textit{\name{}, one checkpoint per model read at every depth}} \\
    \rowcolor{loopblue!6}  &  & 1 & \textbf{15} & 100.0 & 70.8 & 31.5 & 58.0 & 41.7 \\
    \rowcolor{loopblue!6}  &  & 2 & 30 & 100.0 & 80.6 & 40.5 & 65.4 & 51.6 \\
    \rowcolor{loopblue!6}  &  & 3 & 44 & 100.0 & 81.9 & 45.0 & 68.0 & 55.2 \\
    \rowcolor{loopblue!6} \multirow{-4}{*}{\shortstack[l]{\name{}-1.4B\\[-1pt]{\scriptsize\color{black!55}Ouro-1.4B}}} & \multirow{-4}{*}{1.4B} & 4 & 58 & 97.9 & 84.7 & 44.1 & 68.0 & 55.7 \\
    \addlinespace[2pt]
    \rowcolor{loopblue!6}  &  & 1 & 51 & 97.9 & 76.4 & 39.6 & 63.2 & 48.1 \\
    \rowcolor{loopblue!6}  &  & 2 & 101 & 100.0 & 88.9 & 46.8 & 71.0 & 60.3 \\
    \rowcolor{loopblue!6}  &  & 3 & 150 & 100.0 & 88.9 & 55.9 & 75.3 & 66.0 \\
    \rowcolor{loopblue!6} \multirow{-4}{*}{\shortstack[l]{\name{}-2.6B\\[-1pt]{\scriptsize\color{black!55}Ouro-2.6B}}} & \multirow{-4}{*}{2.6B} & 4 & 200 & 100.0 & \textbf{90.3} & \textbf{56.8} & \textbf{76.2} & \textbf{67.3} \\
    \addlinespace[2pt]
    \rowcolor{loopblue!6}  &  & 1 & 23 & 100.0 & 83.3 & 46.8 & 69.3 & 57.1 \\
    \rowcolor{loopblue!6} \multirow{-2}{*}{\shortstack[l]{\name{}-3B\\[-1pt]{\scriptsize\color{black!55}Nanbeige4.2-3B}}} & \multirow{-2}{*}{3B} & 2 & 45 & 100.0 & \textbf{90.3} & 49.5 & 72.7 & 62.8 \\
    \midrule
    \rowcolor{rowgray} \multicolumn{9}{l}{\textit{Dense controls, same data and recipe}} \\
    Qwen3-1.7B & 1.7B & -- & 22 & 100.0 & 84.7 & 34.2 & 63.6 & 50.0 \\
    Qwen3.5-2B & 2B & -- & 105 & 100.0 & 88.9 & 42.3 & 68.8 & 57.5 \\
    Qwen3.5-4B & 4B & -- & 140 & 100.0 & \textbf{90.3} & 51.4 & 73.6 & 63.9 \\
    \midrule
    \rowcolor{rowgray} \multicolumn{9}{l}{\textit{Released references, trained differently}} \\
    Kev-0.8B & 0.8B & -- & 79 & 100.0 & 80.6 & 36.9 & 63.6 & 49.3 \\
    Kev-4B & 4B & -- & 108 & 100.0 & 93.1 & 54.1 & 75.8 & 67.2 \\
    Jev 1.13.0 & n/a & -- & 652$^\dagger$ & 100.0 & 98.6 & 73.0 & 86.6 & 82.2 \\
    \bottomrule
  \end{tabular}
  \caption{Accuracy and latency on the public JevBench items per tier (48 easy, 72 standard, 111 hard), with the JevBench Intelligence score. Bold marks the best value among the models trained on identical data.}
  \label{tab:jevbench}
  \raggedright\footnotesize
  $^\dagger$Measured through the production API with the network included, and not comparable to the local measurements.
\end{table}

\Cref{fig:pareto} places the \name{} checkpoints and the dense models on the 231 public JevBench items (\cref{sec:results}).

\begin{figure}[!t]
  \begin{minipage}[b]{0.5\textwidth}
    \centering
    \includegraphics[width=\linewidth]{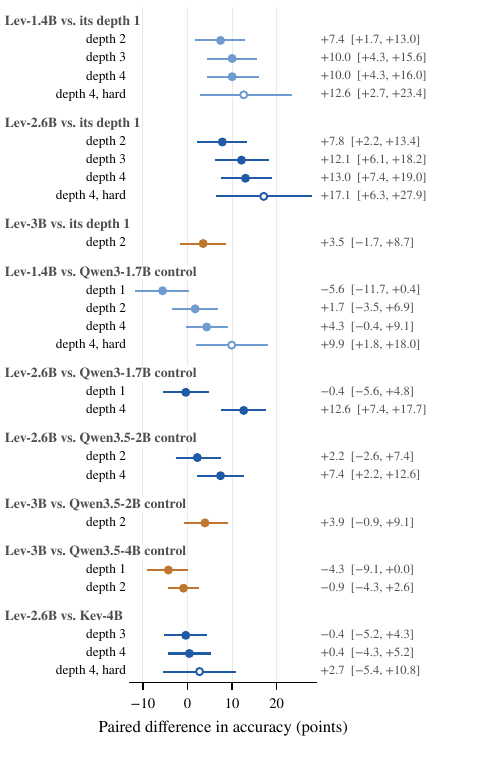}
  \caption{Paired item-level differences in accuracy on the public JevBench items (231 in all, 111 hard), in points with 95\% CIs. Open markers are the hard items.}
  \label{fig:paired}
  \end{minipage}\hfill
  \begin{minipage}[b]{0.47\textwidth}
    \centering
  \centering
  \small
  \renewcommand{\arraystretch}{1.08}
  \begin{tabular}{lccc}
    \toprule
    System & Depth & Hard ECE & Hard acc. \\
    \midrule
    \rowcolor{rowgray} \multicolumn{4}{l}{\textit{\name{}, one checkpoint per model read at every depth}} \\
    \rowcolor{loopblue!6}  & 1 & 0.255 & 31.5 \\
    \rowcolor{loopblue!6}  & 2 & 0.237 & 40.5 \\
    \rowcolor{loopblue!6}  & 3 & 0.226 & 45.0 \\
    \rowcolor{loopblue!6} \multirow{-4}{*}{\name{}-1.4B} & 4 & 0.231 & 44.1 \\
    \addlinespace[2pt]
    \rowcolor{loopblue!6}  & 1 & 0.183 & 39.6 \\
    \rowcolor{loopblue!6}  & 2 & 0.241 & 46.8 \\
    \rowcolor{loopblue!6}  & 3 & 0.202 & 55.9 \\
    \rowcolor{loopblue!6} \multirow{-4}{*}{\name{}-2.6B} & 4 & 0.192 & 56.8 \\
    \addlinespace[2pt]
    \rowcolor{loopblue!6}  & 1 & 0.219 & 46.8 \\
    \rowcolor{loopblue!6} \multirow{-2}{*}{\name{}-3B} & 2 & 0.221 & 49.5 \\
    \midrule
    \rowcolor{rowgray} \multicolumn{4}{l}{\textit{Dense control, same data and recipe}} \\
    Qwen3-1.7B & -- & 0.253 & 34.2 \\
    \midrule
    \rowcolor{rowgray} \multicolumn{4}{l}{\textit{Released references, trained differently}} \\
    Kev-0.8B & -- & 0.172 & 36.9 \\
    Kev-4B & -- & 0.112 & 54.1 \\
    Jev 1.13.0 & -- & 0.086 & 73.0 \\
    \bottomrule
  \end{tabular}
  \captionof{table}{Top-label expected calibration error on the 111 hard JevBench items, with the \name{} readouts and the control at their fitted temperatures.}
  \label{tab:calibration}
  \end{minipage}
\end{figure}

\subsection{Paired Comparisons}

\Cref{fig:paired} shows the paired differences on JevBench behind \cref{sec:results}.
The McNemar test gives $p<0.001$ for depth four against depth one of \name{}-2.6B and $p=0.002$ for \name{}-1.4B.
\name{}-2.6B is ahead of \name{}-1.4B by 8.2 points at depth four (95\% CI $[+3.5, +13.0]$) and by 12.6 points on the hard tier ($[+4.5, +21.6]$).

\subsection{Calibration}

\Cref{tab:calibration} lists the hard-tier ECE of every system, with the \name{} readouts and the control at their fitted temperatures (\cref{sec:analysis:calibration}).
The fitted temperatures lie between 1.05 and 1.18 for the Ouro readouts and are 1.00 and 1.02 for \name{}-3B, so on the calibration partition the raw readouts are at most slightly overconfident.
Temperature scaling lowers the hard-tier ECE at depth four from 0.255 to 0.231 for \name{}-1.4B and from 0.218 to 0.192 for \name{}-2.6B, and that of the Qwen3-1.7B control from 0.258 to 0.253.
After scaling, the ECE of \name{}-1.4B falls from depth one to depth three and rises slightly at depth four, and \name{}-2.6B has its lowest ECE at depth one and highest at depth two.

\section{On-Demand Depth}
\label{app:ondemand}

\Cref{tab:ondemand} in \cref{sec:analysis:ondemand} gives the pooled held-out numbers on development and on the locked test, and this section gives every panel.

\subsection{Development Panels}

\Cref{tab:adaptive} gives on-demand depth on every development panel.
An oracle shows the headroom.
On the tuning panels, stopping each question at the earliest depth that answers it correctly would reach 85.9\% at a cost of 0.34, against 77.4\% for the deepest readout.
The temperature per depth matters only for calibration.
With one temperature for all depths or none, on-demand accuracy on the tuning and held-out panels changes by at most 0.2 points.

\begin{table}[!htbp]
  \centering
  \small
  \renewcommand{\arraystretch}{1.08}
  \setlength{\tabcolsep}{4.5pt}
  \begin{tabular}{llccccccc}
    \toprule
    Model & Panel & Deepest & On-demand & Cost & \multicolumn{2}{c}{vs.\ deepest} & \multicolumn{2}{c}{vs.\ frontier} \\
    \midrule
    \multirow{4}{*}{\name{}-1.4B}
      & \levbench{} & 58.1 & 57.5 & 0.80 & $-0.6$ & \ci{$[-1.0, -0.3]$} & $+0.8$ & \ci{$[+0.1, +1.4]$} \\
      & \kevbreadth{} & 66.0 & 66.1 & 0.75 & $+0.1$ & \ci{$[-0.4, +0.6]$} & $-0.2$ & \ci{$[-1.1, +0.7]$} \\
      & \cellcolor{loopblue!6}held-out, pooled & \cellcolor{loopblue!6}61.3 & \cellcolor{loopblue!6}61.0 & \cellcolor{loopblue!6}0.78 & \cellcolor{loopblue!6}$-0.3$ & \cellcolor{loopblue!6}\ci{$[-0.6, -0.0]$} & \cellcolor{loopblue!6}$+0.5$ & \cellcolor{loopblue!6}\ci{$[-0.0, +1.0]$} \\
      & JevBench & 68.0 & 67.1 & 0.59 & $-0.9$ & \ci{$[-2.6, +0.9]$} & $+0.8$ & \ci{$[-2.2, +4.0]$} \\
    \midrule
    \multirow{8}{*}{\name{}-2.6B}
      & \kevseen{} & 86.1 & 85.9 & 0.44 & $-0.2$ & \ci{$[-1.0, +0.5]$} & $+1.8$ & \ci{$[+0.6, +3.1]$} \\
      & \kevtransfer{} & 72.5 & 71.5 & 0.71 & $-0.9$ & \ci{$[-1.8, -0.2]$} & $+1.0$ & \ci{$[-0.4, +2.5]$} \\
      & \kevskills{} & 71.4 & 70.5 & 0.85 & $-0.9$ & \ci{$[-1.7, -0.2]$} & $-0.3$ & \ci{$[-1.5, +0.7]$} \\
      & \cellcolor{loopblue!6}tuning, pooled & \cellcolor{loopblue!6}77.4 & \cellcolor{loopblue!6}76.7 & \cellcolor{loopblue!6}0.64 & \cellcolor{loopblue!6}$-0.7$ & \cellcolor{loopblue!6}\ci{$[-1.1, -0.2]$} & \cellcolor{loopblue!6}$+1.4$ & \cellcolor{loopblue!6}\ci{$[+0.6, +2.1]$} \\
    \cmidrule(l){2-9}
      & \levbench{} & 63.6 & 62.8 & 0.76 & $-0.9$ & \ci{$[-1.3, -0.5]$} & $+0.1$ & \ci{$[-0.6, +0.9]$} \\
      & \kevbreadth{} & 67.7 & 67.3 & 0.76 & $-0.4$ & \ci{$[-0.9, +0.2]$} & $+0.3$ & \ci{$[-0.6, +1.0]$} \\
      & \cellcolor{loopblue!6}held-out, pooled & \cellcolor{loopblue!6}65.3 & \cellcolor{loopblue!6}64.6 & \cellcolor{loopblue!6}0.76 & \cellcolor{loopblue!6}$-0.7$ & \cellcolor{loopblue!6}\ci{$[-1.0, -0.4]$} & \cellcolor{loopblue!6}$+0.2$ & \cellcolor{loopblue!6}\ci{$[-0.4, +0.8]$} \\
      & JevBench & 76.2 & 74.0 & 0.61 & $-2.2$ & \ci{$[-4.3, -0.4]$} & $+1.2$ & \ci{$[-2.4, +4.4]$} \\
    \midrule
    \multirow{4}{*}{\name{}-3B}
      & \levbench{} & 59.4 & 58.5 & 0.87 & $-0.9$ & \ci{$[-1.4, -0.6]$} & $+1.0$ & \ci{$[+0.5, +1.4]$} \\
      & \kevbreadth{} & 66.1 & 65.4 & 0.82 & $-0.7$ & \ci{$[-1.2, -0.2]$} & $+0.6$ & \ci{$[+0.1, +1.2]$} \\
      & \cellcolor{loopblue!6}held-out, pooled & \cellcolor{loopblue!6}62.1 & \cellcolor{loopblue!6}61.3 & \cellcolor{loopblue!6}0.85 & \cellcolor{loopblue!6}$-0.8$ & \cellcolor{loopblue!6}\ci{$[-1.1, -0.5]$} & \cellcolor{loopblue!6}$+1.0$ & \cellcolor{loopblue!6}\ci{$[+0.6, +1.3]$} \\
      & JevBench & 72.7 & 71.4 & 0.70 & $-1.3$ & \ci{$[-3.5, +0.4]$} & $+0.8$ & \ci{$[-2.2, +3.7]$} \\
    \bottomrule
  \end{tabular}
  \caption{On-demand depth with the stopper of \cref{sec:method:stopper}. Cost is the token-weighted share of the deepest pass. The last four columns are paired differences in accuracy points with 95\% CIs, against the deepest readout and against the fixed-depth frontier at the same cost. Only \name{}-2.6B is read on the tuning panels.}
  \label{tab:adaptive}
\end{table}

\subsection{Locked Test}

\Cref{tab:test} gives on-demand depth on every test partition, read once with the final configuration, and \cref{tab:ondemand} repeats its pooled held-out rows.
The test partitions of the three tuning suites are reported but did not enter any decision.

\begin{table}[!htbp]
  \centering
  \small
  \renewcommand{\arraystretch}{1.08}
  \setlength{\tabcolsep}{4.5pt}
  \begin{tabular}{llccccccc}
    \toprule
    Model & Panel (test) & Deepest & On-demand & Cost & \multicolumn{2}{c}{vs.\ deepest} & \multicolumn{2}{c}{vs.\ frontier} \\
    \midrule
    \multirow{6}{*}{\name{}-1.4B}
      & \levbench{} & 58.4 & 57.5 & 0.81 & $-0.9$ & \ci{$[-1.2, -0.6]$} & $+0.5$ & \ci{$[+0.1, +1.0]$} \\
      & \kevbreadth{} & 64.9 & 64.7 & 0.75 & $-0.3$ & \ci{$[-0.8, +0.2]$} & $+0.3$ & \ci{$[-0.6, +1.2]$} \\
      & \cellcolor{loopblue!6}held-out, pooled & \cellcolor{loopblue!6}60.1 & \cellcolor{loopblue!6}59.3 & \cellcolor{loopblue!6}0.80 & \cellcolor{loopblue!6}$-0.7$ & \cellcolor{loopblue!6}\ci{$[-1.0, -0.5]$} & \cellcolor{loopblue!6}$+0.5$ & \cellcolor{loopblue!6}\ci{$[+0.1, +0.9]$} \\
    \cmidrule(l){2-9}
      & \kevseen{} & 85.3 & 84.0 & 0.50 & $-1.3$ & \ci{$[-2.2, -0.5]$} & $-0.3$ & \ci{$[-1.7, +1.1]$} \\
      & \kevtransfer{} & 70.6 & 68.4 & 0.75 & $-2.2$ & \ci{$[-3.2, -1.4]$} & $+0.2$ & \ci{$[-1.6, +2.0]$} \\
      & \kevskills{} & 70.5 & 69.5 & 0.86 & $-1.0$ & \ci{$[-1.8, -0.3]$} & $-0.5$ & \ci{$[-1.5, +0.5]$} \\
    \midrule
    \multirow{6}{*}{\name{}-2.6B}
      & \levbench{} & 64.4 & 63.6 & 0.78 & $-0.8$ & \ci{$[-1.1, -0.6]$} & $+0.3$ & \ci{$[-0.2, +0.8]$} \\
      & \kevbreadth{} & 65.3 & 64.2 & 0.75 & $-1.0$ & \ci{$[-1.5, -0.6]$} & $-0.5$ & \ci{$[-1.4, +0.3]$} \\
      & \cellcolor{loopblue!6}held-out, pooled & \cellcolor{loopblue!6}64.6 & \cellcolor{loopblue!6}63.7 & \cellcolor{loopblue!6}0.78 & \cellcolor{loopblue!6}$-0.9$ & \cellcolor{loopblue!6}\ci{$[-1.1, -0.6]$} & \cellcolor{loopblue!6}$+0.1$ & \cellcolor{loopblue!6}\ci{$[-0.4, +0.5]$} \\
    \cmidrule(l){2-9}
      & \kevseen{} & 87.3 & 86.2 & 0.44 & $-1.0$ & \ci{$[-1.9, -0.2]$} & $+1.4$ & \ci{$[+0.4, +2.5]$} \\
      & \kevtransfer{} & 75.3 & 74.3 & 0.70 & $-1.0$ & \ci{$[-2.0, -0.2]$} & $+0.6$ & \ci{$[-0.7, +2.0]$} \\
      & \kevskills{} & 73.4 & 72.2 & 0.80 & $-1.2$ & \ci{$[-2.0, -0.4]$} & $-0.5$ & \ci{$[-1.9, +0.9]$} \\
    \midrule
    \multirow{6}{*}{\name{}-3B}
      & \levbench{} & 59.3 & 58.2 & 0.88 & $-1.1$ & \ci{$[-1.4, -0.8]$} & $+0.9$ & \ci{$[+0.6, +1.2]$} \\
      & \kevbreadth{} & 64.8 & 64.2 & 0.82 & $-0.5$ & \ci{$[-1.1, -0.0]$} & $+0.5$ & \ci{$[-0.1, +1.1]$} \\
      & \cellcolor{loopblue!6}held-out, pooled & \cellcolor{loopblue!6}60.7 & \cellcolor{loopblue!6}59.8 & \cellcolor{loopblue!6}0.87 & \cellcolor{loopblue!6}$-1.0$ & \cellcolor{loopblue!6}\ci{$[-1.2, -0.7]$} & \cellcolor{loopblue!6}$+0.9$ & \cellcolor{loopblue!6}\ci{$[+0.6, +1.2]$} \\
    \cmidrule(l){2-9}
      & \kevseen{} & 85.8 & 84.7 & 0.61 & $-1.1$ & \ci{$[-2.1, -0.3]$} & $+0.4$ & \ci{$[-0.6, +1.2]$} \\
      & \kevtransfer{} & 72.2 & 70.8 & 0.78 & $-1.3$ & \ci{$[-2.3, -0.5]$} & $+2.5$ & \ci{$[+1.4, +3.7]$} \\
      & \kevskills{} & 69.5 & 69.2 & 0.89 & $-0.3$ & \ci{$[-1.1, +0.6]$} & $+0.7$ & \ci{$[-0.2, +1.5]$} \\
    \bottomrule
  \end{tabular}
  \caption{On-demand depth on the test partitions, in the format of \cref{tab:adaptive}. The pooled row combines the two held-out panels.}
  \label{tab:test}
\end{table}

\subsection{Further Panels}

\Cref{tab:extra} adds the development partitions of two training sources, consumer-complaint documents (\texttt{documents-v1}, 920 questions) and developer-tooling decisions (\texttt{devtools-v1}, 1,074), and 1,000 MMLU-Pro questions from a public external set that no training set includes.
As on the tiers of \levbench{} (\cref{sec:analysis:tiers}), every model computes less on the two training sources than on MMLU-Pro.

\begin{table}[!htbp]
  \centering
  \small
  \renewcommand{\arraystretch}{1.08}
  \setlength{\tabcolsep}{4.5pt}
  \begin{tabular}{llccccccc}
    \toprule
    Model & Panel & Deepest & On-demand & Cost & \multicolumn{2}{c}{vs.\ deepest} & \multicolumn{2}{c}{vs.\ frontier} \\
    \midrule
    \multirow{3}{*}{\name{}-1.4B}
      & \texttt{documents-v1} & 87.1 & 87.0 & 0.57 & $-0.1$ & \ci{$[-1.0, +0.8]$} & $+0.0$ & \ci{$[-1.2, +1.3]$} \\
      & \texttt{devtools-v1} & 67.0 & 66.4 & 0.67 & $-0.7$ & \ci{$[-1.5, +0.1]$} & $+0.1$ & \ci{$[-1.3, +1.4]$} \\
      & MMLU-Pro & 38.6 & 37.6 & 0.88 & $-1.0$ & \ci{$[-1.8, -0.3]$} & $-0.6$ & \ci{$[-1.7, +0.5]$} \\
    \midrule
    \multirow{3}{*}{\name{}-2.6B}
      & \texttt{documents-v1} & 88.7 & 87.9 & 0.43 & $-0.8$ & \ci{$[-1.7, +0.2]$} & $+2.1$ & \ci{$[+1.0, +3.2]$} \\
      & \texttt{devtools-v1} & 69.2 & 68.1 & 0.74 & $-1.1$ & \ci{$[-1.9, -0.4]$} & $-0.1$ & \ci{$[-1.4, +1.2]$} \\
      & MMLU-Pro & 48.7 & 47.9 & 0.83 & $-0.8$ & \ci{$[-1.6, -0.1]$} & $+0.3$ & \ci{$[-1.0, +1.6]$} \\
    \midrule
    \multirow{3}{*}{\name{}-3B}
      & \texttt{documents-v1} & 88.3 & 87.5 & 0.65 & $-0.8$ & \ci{$[-1.6, +0.0]$} & $+2.3$ & \ci{$[+1.2, +3.5]$} \\
      & \texttt{devtools-v1} & 69.1 & 67.4 & 0.82 & $-1.7$ & \ci{$[-2.8, -0.6]$} & $-0.3$ & \ci{$[-1.2, +0.7]$} \\
      & MMLU-Pro & 46.6 & 46.2 & 0.92 & $-0.4$ & \ci{$[-1.1, +0.3]$} & $+1.1$ & \ci{$[+0.4, +1.9]$} \\
    \bottomrule
  \end{tabular}
  \caption{On-demand depth on further panels, in the format of \cref{tab:adaptive}.}
  \label{tab:extra}
\end{table}

\subsection{Tiers and Difficulty}

\Cref{tab:tiers} splits on-demand depth of \name{}-2.6B by the tiers of \levbench{} (\cref{sec:analysis:tiers}).

\begin{table}[H]
  \centering
  \small
  \renewcommand{\arraystretch}{1.08}
  \setlength{\tabcolsep}{3pt}
  \begin{tabular}{lcccc}
    \toprule
    Tier & Depth 1 / 4 & On-demand & Cost & vs.\ frontier \\
    \midrule
    id & 62.6 / 75.8 & 75.3 & 0.66 & $+1.8$ \ci{$[+0.6, +3.0]$} \\
    near & 42.0 / 71.8 & 69.8 & 0.79 & $-0.3$ \ci{$[-2.4, +1.4]$} \\
    deep & 34.5 / 45.4 & 44.9 & 0.87 & $-0.3$ \ci{$[-1.5, +0.9]$} \\
    far & 41.5 / 57.5 & 56.8 & 0.75 & $-0.2$ \ci{$[-1.6, +1.2]$} \\
    \bottomrule
  \end{tabular}
  \caption{On-demand depth of \name{}-2.6B on the tiers of \levbench{}, ordered by distance from the training data.}
  \label{tab:tiers}
\end{table}

\begin{figure}[!htbp]
  \centering
  \begin{tikzpicture}
    \node[inner sep=0pt, anchor=south west] (img) {\includegraphics[trim=360bp 0 0 0, clip, width=0.32\textwidth]{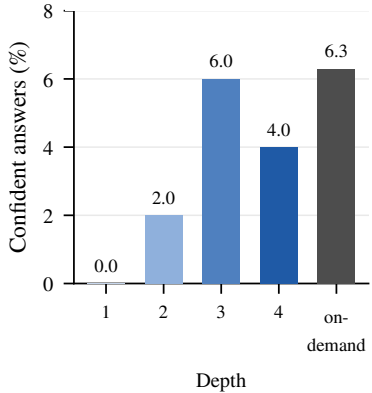}};
    \begin{scope}[x={(img.south east)}, y={(img.north west)}]
      \fill[white] (0.19, 0.88) rectangle (1, 1);
    \end{scope}
  \end{tikzpicture}
  \caption{Share of the 300 unknowable \levbench{} questions that \name{}-2.6B answers with a top probability of at least 0.9, by depth and on demand.}
  \label{fig:unknowable}
\end{figure}

\Cref{fig:difficulty} in \cref{sec:analysis:ondemand} breaks the \levbench{} questions that carry a difficulty $k$ down by $k$, and \cref{fig:unknowable} reports how often \name{}-2.6B is confident on its 300 unknowable questions.

\section{Stopping Rules}
\label{app:stopper}

\subsection{Stopper and Top-Probability Rule}

\paragraph{Stopper.}
Each per-depth stopper is a logistic regression with scikit-learn's default L2 regularisation on the eight features of \cref{sec:method:stopper}, computed from readouts at their fitted temperatures.
At depth one there is no earlier readout, so the features that compare with depth $k-1$ take the values of an unchanged readout.
The threshold is searched over 201 values in $[0.5, 1]$.
The two folds of each panel are assigned by a hash of the record group, so related questions share a fold.
Fitted on the calibration partition alone, the stopper has the threshold 0.915 for both Ouro models and 0.8075 for \name{}-3B.

\paragraph{Top-probability rule.}
A question stops at the first depth whose top probability, after temperature scaling, reaches a threshold.
We choose this threshold on the calibration partition as the cheapest threshold whose accuracy is within one point of the deepest readout, which gives 0.81 for \name{}-2.6B, 0.76 for \name{}-1.4B, and 0.69 for \name{}-3B.

\subsection{Shallow Confidence}

\Cref{tab:diagnostic} gives the diagnosis of \cref{sec:analysis:stop} for \name{}-2.6B on every panel.
The stopped questions are those whose depth-one top probability reaches the threshold of the top-probability rule.
The first AUROC separates right from wrong depth-one answers by the depth-one top probability.
The second does so among the questions that depth one gets wrong, for those that the deepest readout fixes against those it does not.

\begin{table}[H]
  \centering
  \small
  \renewcommand{\arraystretch}{1.08}
  \setlength{\tabcolsep}{3pt}
  \begin{tabular}{lcccc}
    \toprule
    & \multicolumn{2}{c}{Stopped at depth 1} & \multicolumn{2}{c}{AUROC} \\
    \cmidrule(lr){2-3}\cmidrule(lr){4-5}
    Panel & Top prob. & Acc. & Right/wrong & Fixed/not \\
    \midrule
    \color{black!55}Calibration & \color{black!55}95.3 & \color{black!55}94.8 & \color{black!55}0.82 & \color{black!55}0.50 \\
    \kevseen{} & 95.5 & 94.7 & 0.83 & 0.46 \\
    \kevtransfer{} & 93.3 & 78.9 & 0.72 & 0.60 \\
    \kevskills{} & 88.8 & 81.6 & 0.67 & 0.57 \\
    \levbench{} & 93.0 & 88.2 & 0.73 & 0.61 \\
    \kevbreadth{} & 90.0 & 77.1 & 0.70 & 0.49 \\
    \bottomrule
  \end{tabular}
  \caption{Depth-one readout of \name{}-2.6B per panel. The first two columns, in percent, describe the questions that the top-probability rule stops at depth one.}
  \label{tab:diagnostic}
\end{table}

\Cref{tab:difficulty} gives the second AUROC for every checkpoint.
More training data does not raise it. On \kevbreadth{}, it stays between 0.39 and 0.53 for every backbone, training set, and full-weight variant.
The ECE on \kevbreadth{} lies between 0.07 and 0.12 for every readout, and neither more training data nor full-weight fine-tuning lowers it consistently.

\begin{table}[H]
  \centering
  \small
  \renewcommand{\arraystretch}{1.08}
  \setlength{\tabcolsep}{2.5pt}
  \begin{tabular}{lcccc}
    \toprule
    & \multicolumn{2}{c}{AUROC} & \multicolumn{2}{c}{ECE, \kevbreadth{}} \\
    \cmidrule(lr){2-3}\cmidrule(lr){4-5}
    Training set & \levbench{} & \kevbreadth{} & Depth 1 & Deepest \\
    \midrule
    \rowcolor{rowgray} \multicolumn{5}{l}{\textit{\name{}-1.4B}} \\
    \rowcolor{loopblue!6} \quad \trainsmall{} & 0.59 & 0.53 & 0.11 & 0.08 \\
    \quad \trainmid{} & 0.58 & 0.49 & 0.11 & 0.10 \\
    \quad \trainlarge{} & 0.59 & 0.50 & 0.12 & 0.12 \\
    \color{black!55}\quad full & \color{black!55}0.56 & \color{black!55}0.49 & \color{black!55}0.12 & \color{black!55}0.09 \\
    \midrule
    \rowcolor{rowgray} \multicolumn{5}{l}{\textit{\name{}-2.6B}} \\
    \rowcolor{loopblue!6} \quad \trainsmall{} & 0.61 & 0.49 & 0.10 & 0.09 \\
    \quad \trainmid{} & 0.64 & 0.47 & 0.08 & 0.07 \\
    \quad \trainlarge{} & 0.58 & 0.49 & 0.10 & 0.11 \\
    \color{black!55}\quad full & \color{black!55}0.61 & \color{black!55}0.45 & \color{black!55}0.07 & \color{black!55}0.07 \\
    \midrule
    \rowcolor{rowgray} \multicolumn{5}{l}{\textit{\name{}-3B}} \\
    \rowcolor{loopblue!6} \quad \trainsmall{} & 0.54 & 0.43 & 0.12 & 0.10 \\
    \quad \trainmid{} & 0.54 & 0.43 & 0.09 & 0.09 \\
    \quad \trainlarge{} & 0.51 & 0.39 & 0.10 & 0.10 \\
    \color{black!55}\quad full & \color{black!55}0.49 & \color{black!55}0.42 & \color{black!55}0.09 & \color{black!55}0.08 \\
    \bottomrule
  \end{tabular}
  \caption{How well the depth-one top probability finds wrong answers that more steps fix, and the calibration of the shallowest and deepest readouts on \kevbreadth{}. Gray rows (full) train all weights on \trainsmall{}.}
  \label{tab:difficulty}
\end{table}

\subsection{Early-Exit Rules and Stopper Variants}

\Cref{tab:ablation} gives the accuracy and cost behind \cref{fig:ablation} in \cref{sec:analysis:stop}, which compares the stopping rules with the fixed-depth frontier on the pooled held-out panels.

\begin{table}[!htbp]
  \centering
  \small
  \renewcommand{\arraystretch}{1.08}
  \setlength{\tabcolsep}{3pt}
  \begin{tabular}{lccccccccc}
    \toprule
    & \multicolumn{3}{c}{\name{}-1.4B} & \multicolumn{3}{c}{\name{}-2.6B} & \multicolumn{3}{c}{\name{}-3B} \\
    \cmidrule(lr){2-4}\cmidrule(lr){5-7}\cmidrule(lr){8-10}
    Rule & Acc. & Cost & vs.\ frontier & Acc. & Cost & vs.\ frontier & Acc. & Cost & vs.\ frontier \\
    \midrule
    \rowcolor{rowgray} \multicolumn{10}{l}{\textit{Rules from the early-exit literature}} \\
    \quad top probability & 60.1 & 0.79 & $-0.5$ \ci{$[-1.1, +0.1]$} & 64.2 & 0.81 & $-0.4$ \ci{$[-1.0, +0.1]$} & 60.9 & 0.81 & $+1.0$ \ci{$[+0.5, +1.4]$} \\
    \quad margin & 59.6 & 0.75 & $-0.9$ \ci{$[-1.5, -0.1]$} & 63.8 & 0.78 & $-0.7$ \ci{$[-1.3, -0.0]$} & 60.3 & 0.77 & $+0.9$ \ci{$[+0.4, +1.4]$} \\
    \quad entropy & 59.8 & 0.80 & $-0.8$ \ci{$[-1.3, -0.2]$} & 62.8 & 0.76 & $-1.6$ \ci{$[-2.3, -0.9]$} & 58.2 & 0.66 & $+0.1$ \ci{$[-0.4, +0.6]$} \\
    \quad patience, $p=1$ & 58.7 & 0.62 & $-0.1$ \ci{$[-0.5, +0.3]$} & 62.9 & 0.63 & $-0.1$ \ci{$[-0.6, +0.3]$} & -- & -- & -- \\
    \quad patience, $p=2$ & 60.9 & 0.87 & $+0.1$ \ci{$[-0.2, +0.4]$} & 65.0 & 0.88 & $+0.1$ \ci{$[-0.2, +0.4]$} & -- & -- & -- \\
    \midrule
    \rowcolor{rowgray} \multicolumn{10}{l}{\textit{Stopper variants}} \\
    \quad correctness target & 59.7 & 0.75 & $-0.8$ \ci{$[-1.4, -0.1]$} & 64.0 & 0.80 & $-0.5$ \ci{$[-1.1, +0.1]$} & 60.1 & 0.77 & $+0.7$ \ci{$[+0.2, +1.2]$} \\
    \quad calibration only & 60.7 & 0.76 & $+0.2$ \ci{$[-0.3, +0.8]$} & 63.8 & 0.73 & $-0.4$ \ci{$[-1.1, +0.1]$} & 60.2 & 0.77 & $+0.8$ \ci{$[+0.3, +1.3]$} \\
    \quad confidence features only & 61.0 & 0.78 & $+0.4$ \ci{$[-0.1, +1.0]$} & 64.6 & 0.76 & $+0.1$ \ci{$[-0.4, +0.7]$} & 61.3 & 0.85 & $+1.0$ \ci{$[+0.6, +1.3]$} \\
    \quad top probability, label-free & 61.0 & 0.80 & $+0.4$ \ci{$[-0.1, +0.8]$} & 64.6 & 0.78 & $+0.1$ \ci{$[-0.4, +0.7]$} & 61.3 & 0.86 & $+0.8$ \ci{$[+0.5, +1.2]$} \\
    \quad change features only & 60.9 & 0.86 & $+0.1$ \ci{$[-0.2, +0.5]$} & 65.0 & 0.85 & $+0.2$ \ci{$[-0.1, +0.6]$} & 62.1 & 1.00 & $+0.0$ \ci{$[+0.0, +0.0]$} \\
    \midrule
    \rowcolor{loopblue!6} stopper & 61.0 & 0.78 & $+0.5$ \ci{$[-0.0, +1.0]$} & 64.6 & 0.76 & $+0.2$ \ci{$[-0.4, +0.8]$} & 61.3 & 0.85 & $+1.0$ \ci{$[+0.6, +1.3]$} \\
    \bottomrule
  \end{tabular}
  \caption{Stopping rules and stopper variants on the pooled held-out panels, with the accuracy difference in points to the fixed-depth frontier at the same cost, and its 95\% CI. The label-free top-probability row sets one threshold the way the stopper's is set.}
  \label{tab:ablation}
\end{table}

\paragraph{Rules and variants.}
The margin and entropy rules choose their thresholds as the top-probability rule does.
Patience-based exit stops once the answer has not changed for $p$ steps, so it needs at least three steps.
The confidence features are the top probability, the margin, the normalised entropy, the log of the number of options, and the depth.
The change features are the agreement with the previous answer, the KL divergence from the previous readout, the change of the top probability, and the depth.
With two steps there is no earlier answer at depth one, so the change features alone never stop \name{}-3B.
The correctness target predicts whether the answer at the current depth is right and is fitted on the labelled calibration partition, with its threshold chosen as for the top-probability rule.
The label-free top-probability row keeps only the top probability as the stopper's feature, so its threshold on that score is set as the stopper's is, on unlabelled requests for 98\% agreement with the deepest answer.

\paragraph{Results.}
On the tuning panels, the top-probability rule of \name{}-2.6B reaches 75.8\% at a cost of 0.69, 0.2 points below the frontier ($[-1.0, +0.6]$).
The calibration-only stopper reaches 75.6\% at 0.61 ($+0.6$, $[-0.3, +1.4]$), and the stopper 76.7\% at 0.64 ($+1.4$, $[+0.6, +2.1]$).
Against random escalation with the same distribution of stopping depths, the top-probability rule of \name{}-2.6B gains 3.8 points on \kevseen{} ($[+2.6, +5.0]$), 2.5 on \kevtransfer{} ($[+1.0, +4.0]$), and 2.2 on \kevskills{} ($[+1.1, +3.3]$).
For \name{}-3B, whose two steps make the frontier a straight line, every rule that stops some questions at depth one, except the entropy threshold, is above it.

\subsection{Unlabelled Requests}
\label{app:unlabelled}

\Cref{fig:unlabelled}a and b fit the stopper of \name{}-2.6B with only $n$ unlabelled questions from the other fold, drawn five times, and $n=0$ fits it on the calibration partition alone.
With $n=0$ the stopper of \name{}-2.6B stops too early on the held-out panels and falls 1.5 points below the deepest readout, against 0.7 points with all questions of the other fold.
\Cref{fig:unlabelled}c instead fits it on all unlabelled requests of one panel and applies it to another.
A stopper fitted on the training sources (\kevseen{}) falls 0.2 to 1.3 points below the frontier on every other panel, while one fitted on any other panel is at most 0.7 points below it anywhere.

\section{Dense Models and Cascades}
\label{app:cascade}

\Cref{tab:cascade} lists the comparisons behind \cref{fig:cascade} in \cref{sec:analysis:cascade}, with their confidence intervals, and the same comparisons on the development partitions.

\begin{table}[!htbp]
  \centering
  \small
  \renewcommand{\arraystretch}{1.08}
  \setlength{\tabcolsep}{2.5pt}
  \begin{tabular}{llcccc}
    \toprule
    & & \multicolumn{2}{c}{Test} & \multicolumn{2}{c}{Development} \\
    \cmidrule(lr){3-4}\cmidrule(lr){5-6}
    Model & Compared with & \levbench{} & \kevbreadth{} & \levbench{} & \kevbreadth{} \\
    \midrule
    \multirow{2}{*}{\name{}-1.4B} & Qwen3-1.7B $\to$ Qwen3.5-4B & $+7.0$ \ci{$[+5.9, +8.1]$} & $+5.5$ \ci{$[+3.9, +7.0]$} & $+7.0$ \ci{$[+5.5, +8.4]$} & $+6.6$ \ci{$[+4.9, +8.2]$} \\
    & best dense option & $+5.4$ \ci{$[+4.5, +6.3]$} & $+4.5$ \ci{$[+3.1, +5.8]$} & $+5.2$ \ci{$[+3.9, +6.4]$} & $+5.5$ \ci{$[+4.1, +6.9]$} \\
    \midrule
    \multirow{5}{*}{\name{}-2.6B} & Qwen3.5-2B $\to$ Qwen3.5-4B & $+6.2$ \ci{$[+5.2, +7.3]$} & $-1.2$ \ci{$[-2.7, +0.3]$} & $+6.1$ \ci{$[+4.6, +7.7]$} & $+0.6$ \ci{$[-0.8, +2.1]$} \\
    & Qwen3-1.7B $\to$ Qwen3.5-4B & $+1.0$ \ci{$[+0.0, +1.9]$} & $-2.4$ \ci{$[-3.9, -1.0]$} & $+0.8$ \ci{$[-0.6, +2.2]$} & $-0.0$ \ci{$[-1.4, +1.2]$} \\
    & Kev-0.8B $\to$ Kev-4B & $+2.1$ \ci{$[+1.0, +3.2]$} & $-1.4$ \ci{$[-3.0, +0.2]$} & $+2.7$ \ci{$[+1.3, +4.1]$} & $+0.8$ \ci{$[-0.7, +2.2]$} \\
    & \name{}-1.4B (depth 4) $\to$ \name{}-2.6B & $+1.2$ \ci{$[+0.7, +1.8]$} & $-1.6$ \ci{$[-2.3, -0.8]$} & $+1.3$ \ci{$[+0.5, +2.1]$} & $-0.3$ \ci{$[-1.1, +0.5]$} \\
    & best dense option & $-0.3$ \ci{$[-1.3, +0.7]$} & $-2.6$ \ci{$[-4.1, -1.1]$} & $-0.6$ \ci{$[-2.0, +0.8]$} & $-0.5$ \ci{$[-1.9, +0.8]$} \\
    \midrule
    \multirow{2}{*}{\name{}-3B} & Qwen3-1.7B $\to$ Qwen3.5-4B & $+8.8$ \ci{$[+7.7, +9.9]$} & $+5.3$ \ci{$[+3.6, +6.9]$} & $+8.6$ \ci{$[+7.1, +10.1]$} & $+6.2$ \ci{$[+4.4, +8.0]$} \\
    & best dense option & $+7.5$ \ci{$[+6.5, +8.4]$} & $+4.6$ \ci{$[+3.2, +6.1]$} & $+7.3$ \ci{$[+6.0, +8.6]$} & $+5.4$ \ci{$[+3.9, +7.0]$} \\
    \midrule
    \rowcolor{rowgray} \multicolumn{6}{l}{\textit{Best dense option, at equal floating-point operations}} \\
    \multicolumn{2}{l}{\quad \name{}-1.4B} & -- & -- & $-6.0$ \ci{$[-7.4, -4.6]$} & $-1.8$ \ci{$[-3.2, -0.3]$} \\
    \multicolumn{2}{l}{\quad \name{}-2.6B} & -- & -- & $-0.7$ \ci{$[-2.1, +0.7]$} & $-0.6$ \ci{$[-1.9, +0.8]$} \\
    \multicolumn{2}{l}{\quad \name{}-3B} & -- & -- & $-5.0$ \ci{$[-6.4, -3.6]$} & $-2.5$ \ci{$[-3.9, -1.1]$} \\
    \bottomrule
  \end{tabular}
  \caption{Accuracy of on-demand \name{} minus a cascade or the best dense option at the same mean latency, in points with 95\% CIs. On the \levbench{} test partition, on-demand \name{}-1.4B, \name{}-2.6B, and \name{}-3B take 46, 155, and 39\,ms on average. A cascade that cannot reach a model's latency is omitted. The last rows compare at equal floating-point operations instead of latency.}
  \label{tab:cascade}
\end{table}

Latencies come from one H200 GPU in bf16, one request at a time, without fused kernels or graph capture.
For each system and depth we time 600 development records three times, take the median, and fit latency as a linear function of the token count, which then prices every request.
An on-demand request pays the latency of the deepest step its state runs.
A cascade escalates with the top-probability rule of \cref{sec:analysis:stop} applied to the small model, and an escalated request pays for both models.
A stopper of the same kind as \name{}'s, which predicts whether the small and the large model agree, leads to the same conclusions.
We compare at the mean latency of on-demand \name{} and mix the two neighbouring operating points of a cascade to reach it.
The best dense option is the upper envelope of every dense model and every dense cascade.
Counted in floating-point operations, a request costs two operations per parameter, token, and loop step.

\section{Training Choices}
\label{app:scale}

\subsection{Training Data and Full Weights}

\Cref{tab:scale} gives the results on the pooled held-out panels, and \cref{tab:scale-panels} splits them by panel.
The full-weight variant uses a ten times lower learning rate for the backbone than LoRA and keeps the same data, schedule, and number of steps.

\begin{table}[!htbp]
  \centering
  \footnotesize
  \renewcommand{\arraystretch}{1.08}
  \setlength{\tabcolsep}{1.8pt}
  \begin{tabular}{llccccccccc}
    \toprule
    & & \multicolumn{4}{c}{\levbench{}} & \multicolumn{4}{c}{\kevbreadth{}} & JevBench \\
    \cmidrule(lr){3-6}\cmidrule(lr){7-10}\cmidrule(lr){11-11}
    Model & Training set & Deepest & On-demand & Cost & vs.\ frontier & Deepest & On-demand & Cost & vs.\ frontier & Deepest \\
    \midrule
\rowcolor{loopblue!6} \name{}-1.4B & \trainsmall{} & 58.1 & 57.5 & 0.80 & $+0.8$ \ci{$[+0.1, +1.4]$} & 66.0 & 66.1 & 0.75 & $-0.2$ \ci{$[-1.1, +0.7]$} & 68.0 \\
 & \trainmid{} & 70.8 & 70.0 & 0.67 & $+0.5$ \ci{$[-0.2, +1.2]$} & 66.7 & 65.9 & 0.75 & $+0.1$ \ci{$[-0.7, +1.0]$} & 69.7 \\
 & \trainlarge{} & 73.0 & 71.9 & 0.61 & $-0.4$ \ci{$[-1.1, +0.3]$} & 65.2 & 64.8 & 0.75 & $-0.3$ \ci{$[-1.2, +0.5]$} & 68.0 \\
 & \color{black!55}full & \color{black!55}57.9 & \color{black!55}56.8 & \color{black!55}0.73 & \color{black!55}$+0.1$ \ci{$[-0.7, +0.8]$} & \color{black!55}63.0 & \color{black!55}63.0 & \color{black!55}0.74 & \color{black!55}$-0.4$ \ci{$[-1.1, +0.4]$} & \color{black!55}68.8 \\
    \midrule
\rowcolor{loopblue!6} \name{}-2.6B & \trainsmall{} & 63.6 & 62.8 & 0.76 & $+0.1$ \ci{$[-0.6, +0.9]$} & 67.7 & 67.3 & 0.76 & $+0.3$ \ci{$[-0.6, +1.0]$} & 76.2 \\
 & \trainmid{} & 73.8 & 72.7 & 0.66 & $+0.4$ \ci{$[-0.3, +1.2]$} & 67.3 & 67.1 & 0.73 & $+0.6$ \ci{$[-0.2, +1.4]$} & 75.3 \\
 & \trainlarge{} & 77.0 & 76.1 & 0.59 & $+1.5$ \ci{$[+0.7, +2.3]$} & 67.3 & 66.6 & 0.70 & $-0.3$ \ci{$[-1.0, +0.4]$} & 77.9 \\
 & \color{black!55}full & \color{black!55}61.6 & \color{black!55}60.8 & \color{black!55}0.76 & \color{black!55}$-0.7$ \ci{$[-1.4, -0.0]$} & \color{black!55}66.0 & \color{black!55}65.6 & \color{black!55}0.67 & \color{black!55}$-0.1$ \ci{$[-0.8, +0.6]$} & \color{black!55}76.6 \\
    \midrule
\rowcolor{loopblue!6} \name{}-3B & \trainsmall{} & 59.4 & 58.5 & 0.87 & $+1.0$ \ci{$[+0.5, +1.4]$} & 66.1 & 65.4 & 0.82 & $+0.6$ \ci{$[+0.1, +1.2]$} & 72.7 \\
 & \trainmid{} & 71.2 & 70.5 & 0.79 & $+1.9$ \ci{$[+1.4, +2.5]$} & 66.5 & 66.5 & 0.81 & $+0.7$ \ci{$[+0.1, +1.2]$} & 74.0 \\
 & \trainlarge{} & 73.6 & 72.6 & 0.74 & $+2.2$ \ci{$[+1.6, +2.8]$} & 66.5 & 66.4 & 0.81 & $+0.9$ \ci{$[+0.3, +1.5]$} & 74.9 \\
 & \color{black!55}full & \color{black!55}59.7 & \color{black!55}58.9 & \color{black!55}0.83 & \color{black!55}$+1.1$ \ci{$[+0.6, +1.5]$} & \color{black!55}66.1 & \color{black!55}65.9 & \color{black!55}0.79 & \color{black!55}$+0.9$ \ci{$[+0.4, +1.5]$} & \color{black!55}74.9 \\
    \bottomrule
  \end{tabular}
  \caption{\Cref{tab:scale} per panel, with the deepest JevBench accuracy of each checkpoint. Gray rows (full) train all weights on \trainsmall{}.}
  \label{tab:scale-panels}
\end{table}

\begin{table}[H]
  \centering
  \small
  \renewcommand{\arraystretch}{1.08}
  \setlength{\tabcolsep}{3pt}
  \begin{tabular}{lcccc}
    \toprule
    Training set & Depth 1 & Deepest & Cost & vs.\ frontier \\
    \midrule
    \rowcolor{rowgray} \multicolumn{5}{l}{\textit{\name{}-1.4B}} \\
    \rowcolor{loopblue!6} \quad \trainsmall{} & 49.6 & 61.3 & 0.78 & $+0.5$ \ci{$[-0.0, +1.0]$} \\
    \quad \trainmid{} & 60.1 & 69.1 & 0.70 & $+0.3$ \ci{$[-0.2, +0.8]$} \\
    \quad \trainlarge{} & 61.3 & 69.8 & 0.67 & $-0.4$ \ci{$[-0.9, +0.1]$} \\
    \color{black!55}\quad full & \color{black!55}50.7 & \color{black!55}60.0 & \color{black!55}0.73 & \color{black!55}$-0.1$ \ci{$[-0.7, +0.4]$} \\
    \midrule
    \rowcolor{rowgray} \multicolumn{5}{l}{\textit{\name{}-2.6B}} \\
    \rowcolor{loopblue!6} \quad \trainsmall{} & 50.5 & 65.3 & 0.76 & $+0.2$ \ci{$[-0.4, +0.8]$} \\
    \quad \trainmid{} & 60.6 & 71.2 & 0.69 & $+0.4$ \ci{$[-0.1, +1.0]$} \\
    \quad \trainlarge{} & 62.9 & 73.0 & 0.63 & $+0.7$ \ci{$[+0.1, +1.2]$} \\
    \color{black!55}\quad full & \color{black!55}53.7 & \color{black!55}63.4 & \color{black!55}0.72 & \color{black!55}$-0.3$ \ci{$[-0.8, +0.1]$} \\
    \midrule
    \rowcolor{rowgray} \multicolumn{5}{l}{\textit{\name{}-3B}} \\
    \rowcolor{loopblue!6} \quad \trainsmall{} & 56.2 & 62.1 & 0.85 & $+1.0$ \ci{$[+0.6, +1.3]$} \\
    \quad \trainmid{} & 64.9 & 69.3 & 0.80 & $+1.4$ \ci{$[+0.9, +1.8]$} \\
    \quad \trainlarge{} & 66.0 & 70.7 & 0.77 & $+1.5$ \ci{$[+1.1, +2.0]$} \\
    \color{black!55}\quad full & \color{black!55}57.9 & \color{black!55}62.3 & \color{black!55}0.81 & \color{black!55}$+1.1$ \ci{$[+0.7, +1.4]$} \\
    \bottomrule
  \end{tabular}
  \caption{Training data and full-weight fine-tuning, on the pooled held-out panels. Cost and the difference to the frontier are for on-demand depth. Gray rows (full) train all weights on \trainsmall{}. Part of \levbench{} is in distribution for \trainmid{} and \trainlarge{} (\cref{app:levbench}).}
  \label{tab:scale}
\end{table}

\subsection{Dense Controls}

We train the dense controls on the three training sets with the same recipe and compare them in \cref{tab:dense-scale} with the deepest readout of the looped model of about the same size.

\begin{table}[H]
  \centering
  \small
  \renewcommand{\arraystretch}{1.08}
  \setlength{\tabcolsep}{5pt}
  \begin{tabular}{lcccccc}
    \toprule
    & \multicolumn{3}{c}{\levbench{}} & \multicolumn{3}{c}{\kevbreadth{}} \\
    \cmidrule(lr){2-4}\cmidrule(lr){5-7}
    Model & 31k & 60k & 120k & 31k & 60k & 120k \\
    \midrule
    \rowcolor{loopblue!6} \name{}-2.6B & 63.6 & 73.8 & 77.0 & 67.7 & 67.3 & 67.3 \\
    Qwen3.5-2B & 53.1 & 66.3 & 69.9 & 64.7 & 64.3 & 65.1 \\
    \midrule
    \rowcolor{loopblue!6} \name{}-3B & 59.4 & 71.2 & 73.6 & 66.1 & 66.5 & 66.5 \\
    Qwen3.5-4B & 62.1 & 71.8 & 74.2 & 67.4 & 69.4 & 69.4 \\
    \bottomrule
  \end{tabular}
  \caption{Accuracy of the deepest readout of the looped models and of dense controls trained on \trainsmall{}, \trainmid{}, and \trainlarge{} (columns). Part of \levbench{} is in distribution for \trainmid{} and \trainlarge{}.}
  \label{tab:dense-scale}
\end{table}

\subsection{Joint Training}
\label{app:joint}

\Cref{tab:joint} gives the numbers behind \cref{fig:training}a in \cref{sec:analysis:joint}, with the paired difference between joint and single-depth training.

\begin{table}[!htbp]
  \centering
  \small
  \renewcommand{\arraystretch}{1.08}
  \setlength{\tabcolsep}{4.5pt}
  \begin{tabular}{ccccccc}
    \toprule
    & \multicolumn{3}{c}{\levbench{}} & \multicolumn{3}{c}{\kevbreadth{}} \\
    \cmidrule(lr){2-4}\cmidrule(lr){5-7}
    Depth & Joint & Single & $\Delta$ & Joint & Single & $\Delta$ \\
    \midrule
    1 & 46.9 & 48.2 & $-1.3$ \ci{$[-2.5, +0.0]$} & 55.7 & 56.9 & $-1.2$ \ci{$[-2.6, +0.3]$} \\
    2 & 58.6 & 58.3 & $+0.4$ \ci{$[-0.6, +1.4]$} & 65.3 & 64.7 & $+0.6$ \ci{$[-0.4, +1.7]$} \\
    4 & 63.6 & 63.2 & $+0.4$ \ci{$[-0.6, +1.4]$} & 67.7 & 67.9 & $-0.2$ \ci{$[-1.4, +0.9]$} \\
    \bottomrule
  \end{tabular}
  \caption{Accuracy of the jointly trained \name{}-2.6B and of a model trained for each depth alone (Single), with the paired difference in points. On JevBench the single-depth models reach 60.6\%, 72.3\%, and 77.9\%, against 63.2\%, 71.0\%, and 76.2\% for joint training.}
  \label{tab:joint}
\end{table}

\paragraph{Training only the last step.}
The model that runs four steps and is trained only at the last leaves its shallow readouts unusable (\cref{sec:analysis:joint}).
Its depth one answers 36.4\% of \levbench{} correctly against 46.9\% for joint training, so its stopper almost never stops there and on-demand depth computes 84\% of the deepest pass instead of 76\%.
It is the single-depth model for depth four of \cref{tab:joint}.

\subsection{Losses and Loop Steps}

\Cref{tab:recipe} lists the recipe variants of \cref{sec:analysis:scale} on the pooled held-out panels.
A Brier term with weight one, added to every step's loss, raises the \levbench{} accuracy of the deepest readout of \name{}-2.6B by 1.3 points but not its \kevbreadth{} accuracy, and has no effect on \name{}-3B.
Weights that grow with the step, $w_t \propto t$, move training toward the deep readouts, so depth one loses 2.2 points on \levbench{} while the deepest readout stays level.
\name{}-2.6B trained and read with six steps gains at most one point after the fourth step on \levbench{} and none on \kevbreadth{}. Its deepest readout is its sixth step, and its depth four reaches 65.9\%.
\name{}-3B trained and read with three steps gains 0.5 points from its third step on the pooled held-out panels.
A second epoch of \name{}-1.4B at half the peak learning rate, started from the first, stays within one point of the first epoch on JevBench at every depth.

\begin{table}[!htbp]
  \centering
  \small
  \renewcommand{\arraystretch}{1.08}
  \setlength{\tabcolsep}{4.5pt}
  \begin{tabular}{lccccc}
    \toprule
    Recipe & Depth 1 & Deepest & On-demand & Cost & vs.\ frontier \\
    \midrule
    \rowcolor{rowgray} \multicolumn{6}{l}{\textit{\name{}-2.6B}} \\
    \rowcolor{loopblue!6} \quad default & 50.5 & 65.3 & 64.6 & 0.76 & $+0.2$ \ci{$[-0.4, +0.8]$} \\
    \quad Brier term & 51.2 & 65.9 & 65.2 & 0.78 & $+0.2$ \ci{$[-0.4, +0.7]$} \\
    \quad $w_t \propto t$ & 48.6 & 65.8 & 65.0 & 0.78 & $+0.3$ \ci{$[-0.2, +0.9]$} \\
    \quad last step only & 38.4 & 65.1 & 64.4 & 0.82 & $+1.1$ \ci{$[+0.6, +1.6]$} \\
    \quad six steps & 50.0 & 66.2 & 65.7 & 0.67 & $-0.2$ \ci{$[-0.7, +0.4]$} \\
    \midrule
    \rowcolor{rowgray} \multicolumn{6}{l}{\textit{\name{}-3B}} \\
    \rowcolor{loopblue!6} \quad default & 56.2 & 62.1 & 61.3 & 0.85 & $+1.0$ \ci{$[+0.6, +1.3]$} \\
    \quad Brier term & 56.4 & 61.6 & 61.0 & 0.85 & $+0.9$ \ci{$[+0.5, +1.2]$} \\
    \quad three steps & 55.7 & 62.7 & 62.1 & 0.66 & $+0.2$ \ci{$[-0.3, +0.6]$} \\
    \bottomrule
  \end{tabular}
  \caption{Recipe variants on the pooled held-out panels. Cost and the difference to the frontier are for on-demand depth, and cost is a share of each model's own deepest pass.}
  \label{tab:recipe}
\end{table}